\documentclass[11pt,a4paper]{article}

\usepackage[comma]{natbib}
\usepackage{color}
\usepackage{graphicx}
\usepackage{pdfsync}

\usepackage{amsmath,amssymb,amsthm,amsopn}
\usepackage[ruled]{algorithm2e}

\usepackage[mathscr]{euscript}

\usepackage{pgfplots}

\pgfplotsset{compat=1.8}

\theoremstyle{remark}

\renewcommand{\today}{\begingroup
\number \day\space  \ifcase \month \or January\or February\or
March\or April\or May\or June\or July\or August\or September\or
October\or November\or December\fi \space  \number \year \endgroup}

\theoremstyle{plain}

\newcommand{\p}{\mathcal P}

\newtheorem{teor*}{Teorema}

\theoremstyle{definition}

\title{A close-up comparison of the misclassification error distance and the adjusted Rand index for external clustering evaluation}

\author{Jos\'e E. Chac\'on\footnote{Departamento de
Matem\'aticas, Universidad de Extremadura, E-06006 Badajoz, Spain. E-mail:
{\tt jechacon@unex.es}}}

\begin{document}

\maketitle

\begin{abstract}
\noindent The misclassification error distance and the adjusted Rand index are two of the most commonly used criteria to evaluate the performance of clustering algorithms. This paper provides an in-depth comparison of the two criteria, aimed to better understand exactly what they measure, their properties and their differences. Starting from their population origins, the investigation includes many data analysis examples and the study of particular cases in great detail. An exhaustive simulation study allows inspecting the criteria distributions and reveals some previous misconceptions.
\end{abstract}


\newpage

\section{Introduction}

The adjusted Rand index (ARI) introduced in \cite{HA85} is one of the most commonly used measures of performance for clustering evaluation. Indeed, it was the recommended choice in the seminal paper of \cite{MC86}, where five criteria were examined regarding the task of comparison of hierarchical clustering algorithms across different hierarchy levels. Their recommendation is based on the fact that, for the null case data (i.e., for a synthetic sample with randomly assigned class labels, showing no significant cluster structure), the ARI was the only index that produced a flat response curve across hierarchy levels, with mean values close to zero, hence indicating that the agreement between the randomly assigned labels and the algorithm solution was due to chance.

Another popular measure for clustering validation, not included in Milligan and Cooper's study, is the misclassification error distance (MED). Its first appearance in the literature dates back at least to \cite{R65}, where it was introduced as a distance between partitions of a finite set, and it was called transfer distance. It is also referred to as partition distance \citep{G02} or maximum matching distance \citep{R15}. Many papers concerning clustering evaluation indeed contain detailed comparisons of both, the ARI and the MED, showing arguments in favour of one or the other; see, for instance, \cite{St03,St04}, \cite{DG06} or \cite{M05,M07,M16}.

Whereas \cite{St04} supports Milligan and Cooper's recommendation by inspecting the performance of the ARI and the MED on an exhaustive simulation study, \cite{M16} suggests that the MED ``comes closest to satisfying everyone'' in terms of its properties and ease of interpretation, \cite{DG06} suggest that the MED is much appropriate for small sample sizes from their study of all the clusterings at a close number of transfers from a given one, and \cite{vL10} considers the MED as ``the most convenient choice from a theoretical point of view''.

It must be stressed that both criteria are commonly categorized as ``external'', in the sense that they are used to measure the performance of a data-based clustering algorithm against a true cluster structure, known in advance in a simulation scenario or after a data inspection by an expert, which is taken as the ideal clustering solution, but is external to the clustering methodology itself. Internal criteria (such as those based on cohesion, entropy, cluster separation, etc) are also frequently used, but they will not constitute the focus of this paper; see \citet{H19} for a thorough review of internal cluster validation indexes.



This paper aims to provide further comparisons between the MED and the ARI, at several levels. Indeed, many other external criteria could be considered as well, and they are also reviewed in the aforementioned comparative studies, but here the discussion is restricted to the former two because they are usually recognized as the main criteria used in practice. The close-up inspection examines a wide range of features: Section \ref{sec:2} first glances through their population origins (i.e., their counterparts in the case where the true underlying data distribution is fully known) and then elaborates on their traditional, and more common, data-based versions. The comparison of these empirical analogues is the subject of Sections \ref{sec:4} (theoretically) and \ref{sec:5} (by simulations). The theoretical study comprises their computation, some illustrations
by means of simple examples, and an analysis of their extreme values in relation to the case of independent clusterings. 
The simulation scenarios investigate the distributions of the criteria in the null case and how they evolve as two clusterings become apart from perfect agreement. Finally, Section \ref{sec:6} discusses the new findings and their implications.

\section{Population and empirical distances between clusterings}\label{sec:2}

\subsection{The population version of cluster analysis}

Cluster analysis is mostly posed as a sample problem, and perhaps that is one of the reasons why many authors have called attention to the lack of theoretical results for clustering \citep{M96,vLBD05}, as opposed to regression or classification, where the population background is much more clearly established.

Traditionally, the goal of clustering techniques is to provide a partitioning of a data set into groups. For that goal, it suffices to have an algorithm which is appropriate for the data set at hand. However, from a statistical perspective, such a given data set is not simply a set of points in the space, but a sample from some probability distribution $P$. Hence, the goal of clustering methodologies can not reduce to partitioning only the data set at hand, but it must provide a mechanism to assign group labels to any point in the space; or, at least, to all the points in the sample space, since they could have been equally drawn as sample points. Such a view of clustering is shared by many authors, including \citet[][p. 245]{Gal02}, \cite{Bal06}, \citet[][p. 196]{K09}, \citet{Ch15} or \citet[][Section 2.3]{W18}.

Hence, the object that clustering algorithms should produce is not only a partition of the data set, but a whole-space partition. This means that if $\Omega$ denotes the sample space, a whole-space clustering is a class of sets $\mathscr C=\{C_1,\dots,C_r\}$ such that $C_i\cap C_j=\varnothing$ for all $i\neq j$ and $C_1\cup\cdots\cup C_r=\Omega$. Indeed, most existing clustering methodologies are able to produce this type of object; this is the case, for instance, for $K$-means clustering, modal clustering or mixture model clustering \citep[see][]{Ch15}. Obviously, any partition of $\Omega$ induces a partition of the observed data set as well. To avoid confusions, these are referred to as a whole-space clustering and a clustering of the data, respectively. Also, note that both objects can have a population version (the partition that would be made if the true underlying distribution were fully known) and a data-based version (the partition that would be made after observing the data).

That made clear, to evaluate the performance of clustering methods from a statistical point of view it is necessary to employ a distance between whole-space clusterings. While there exist many notions of distance between partitions of a finite set \citep{D81,M16}, proposals to serve as a distance between whole-space clusterings do not abound in the literature. Two of them are described next.

First, since the parts of a clustering (i.e., the clusters) are sets, it seems natural for distances between clusterings to be built upon a notion of discrepancy between sets. A usual way to express the discrepancy between two sets $C$ and $D$ is by quantifying the content of their symmetric difference $C\triangle D$. This difference is defined as the elements that $C$ and $D$ do not have in common; that is, $C\triangle D=(C\cup D)\setminus(C\cap D)$. Then, taking into account the distinctive features of a partition, this natural distance between sets can be extended to define a distance between two clusterings $\mathscr C=\{C_1,\dots,C_r\}$ and $\mathscr D=\{D_1,\dots,D_s\}$, by adding up the contributions of the regions that their most similar clusters do not have in common. Specifically, \cite{Ch15} defined the distance in measure between $\mathscr C$ and $\mathscr D$ as
\begin{equation}\label{eq:DIM}
d_{\rm M}(\mathscr C,\mathscr D)=\frac12\min_{\sigma\in \mathcal P_s}\sum_{i=1}^sP(C_i\triangle D_{\sigma(i)}),
\end{equation}
where $\mathcal P_s$ is the set of all permutations of $s$ elements and, without loss of generality, it is assumed that $r\leq s$ so that $\mathscr C$ would be enlarged by adding $s-r$ empty sets $C_{r+1}=\cdots=C_s=\varnothing$ if necessary. More intuitively, $d_{\rm M}(\mathscr C,\mathscr D)$ represents the minimum probability mass that needs to be moved (or re-labeled) to transform $\mathscr C$ into $\mathscr D$, or viceversa.

The above $d_{\rm M}(\mathscr C,\mathscr D)$ is a clustering distance, in the sense of \citet[][Definition 3]{Bal06}. Nevertheless, these authors considered a different distance between whole-space clusterings, $d_{\rm H}(\mathscr C,\mathscr D)$, which they called Hamming distance. This second distance is more closely related to the Rand index (as detailed below), since it is defined as the probability that two independent random observations (drawn from $P$) belong to the same cluster with respect to one of the clusterings and to different clusters with respect to the other clustering. Hence, it can be shown that an explicit expression for this Hamming distance is
\begin{equation}\label{eq:Hamming}
d_{\rm H}(\mathscr C,\mathscr D)=\sum_{i=1}^rP^2(C_i)+\sum_{j=1}^sP^2(D_j)-2\sum_{i=1}^r\sum_{j=1}^sP^2(C_i\cap D_j).
\end{equation}
The dependence of this measure on squared probabilities may appear somehow unnatural, but it is a consequence of the fact that it is based on comparing the cluster labels of pairs of points.

\subsection{Comparing two clusterings of the data}

In a simulation setting, where the true underlying distribution $P$ is fully known, it is possible to compute the ideal population clustering; that is, the whole-space partition that would be made on the basis of this knowledge of $P$ (this ideal partition varies from one methodology to another, depending on the notion of cluster that they seek after). Hence, it is natural to evaluate the performance of a clustering technique by means of the distance from the produced data-based clustering to its population counterpart. Since both are clusterings of the whole space, any of the previously mentioned distances between whole-space clusterings can be employed.

Of course, things are different when dealing with real data. Suppose that we have observed $n$ data points $\mathcal X=\{x_1,\dots,x_n\}$. Even if the usual methods are able to produce whole-space clusterings with the sole information provided by $\mathcal X$, the fact that a clustering distance depends on $P$ \citep{Bal06}, which is unknown for real data sets, implies that to compute the clustering distance in practice it is necessary to replace $P$ by the empirical distribution $P_n$, which assigns probability mass $1/n$ to each data point. This means that only the labels of the data points are used in the comparison between the two clusterings, so that a distance between whole-space clusterings becomes in fact a distance between two clusterings of the data.

When this reasoning is applied to the two distances in the previous section, it results in two well-known distances between partitions of a finite set. To see this, given two partitions $\mathscr C=\{C_1,\dots,C_r\}$ and $\mathscr D=\{D_1,\dots,D_s\}$ of $\mathcal X$, with $r\leq s$, denote by $n_{ij}$,  $n_{i+}=\sum_{j=1}^sn_{ij}$ and $n_{+j}=\sum_{i=1}^rn_{ij}$ the cardinalities of $C_i\cap D_j$, $C_i$ and $D_j$, respectively. The $(r\times s)$-matrix $\mathbf N=(n_{ij})$ is known as the confusion matrix (or contingency table), and the vectors $(n_{1+},\dots,n_{r+})$ and $(n_{+1},\dots,n_{+s})$ constitute its row-wise and column-wise margins, respectively. Then, taking into account that $P(C_i\triangle D_j)=P(C_i)+P(D_j)-2P(C_i\cap D_j)$, it follows that the empirical version of the distance in measure (\ref{eq:DIM}) is
$$\widehat d_{\rm M}(\mathscr C,\mathscr D)=\frac12\min_{\sigma\in \mathcal P_s}n^{-1}\sum_{i=1}^s\big\{n_{i+}+n_{+\sigma(i)}-2n_{i,\sigma(i)}\big\}=1-n^{-1}\max_{\sigma\in\mathcal P_s}\sum_{i=1}^rn_{i,\sigma(i)},$$
which coincides with the definition of the misclassification error distance \citep[see][]{M05}, so that it will be denoted as ${\rm MED}(\mathscr C,\mathscr D)=\widehat d_{\rm M}(\mathscr C,\mathscr D)$ henceforth (or simply MED, if it is obvious which clusterings are being compared). The MED inherits from its population version a clear interpretation as the minimum proportion of data points that would need to be re-labeled so that $\mathscr C$ and $\mathscr D$ coincided, and that is why it is also known as transfer distance \citep{R65}.

On the other hand, the empirical equivalent of the Hamming distance (\ref{eq:Hamming}) is
\begin{equation}\label{eq:empHam}
\widehat d_{\rm H}(\mathscr C,\mathscr D)=n^{-2}\Big\{\sum_{i=1}^rn_{i+}^2+\sum_{j=1}^sn_{+j}^2-2\sum_{i=1}^r\sum_{j=1}^sn_{ij}^2\Big\},
\end{equation}
which is also known as equivalence mismatch coefficient (\citealp{MC70}; \citealp[][p. 241]{Mir96}) or as $n$-invariant Mirkin metric \citep{M16}. Being a sample equivalent of (\ref{eq:Hamming}), $\widehat d_{\rm H}(\mathscr C,\mathscr D)$ equals the proportion of pairs $\{(x_k,x_l)\colon k,l=1,\dots,n\}$ that belong to the same cluster in one of the clusterings and to different clusters in the other clustering. Note that, somehow artificially,
this empirical version of the Hamming distance is taking into account data pairs of type $(x_k,x_k)$ as well.

In statistical terms, if $X,Y$ are independent random variables with distribution $P$ and we denote by $I_A$ the indicator function of a set $A$, the squared probability $P^2(C_i)=E[I_{C_i}(X)I_{C_i}(Y)]$ appearing in (\ref{eq:Hamming}) is estimated in (\ref{eq:empHam}) by the observed value of the $V$-statistic $n^{-2}\sum_{k,l=1}^nI_{C_i}(x_k)I_{C_i}(x_l)=n_{i+}^2/n^2=P_n^2(C_i)$. However, $U$-statistics theory \citep{L90} shows that a better estimate of $P^2(C_i)$ is ${n\choose 2}^{-1}\sum_{1\leq k<l\leq n}I_{C_i}(x_k)I_{C_i}(x_l)={n\choose 2}^{-1}{n_{i+}\choose 2}$. Reasoning similarly for the other terms in (\ref{eq:Hamming}) and making these changes everywhere in (\ref{eq:empHam}) yields the definition of the Rand distance
\begin{equation}\label{eq:RD}
{\rm RD}(\mathscr C,\mathscr D)={n\choose 2}^{-1}\left\{\sum_{i=1}^r{n_{i+}\choose 2}+\sum_{j=1}^s{n_{+j}\choose 2}-2\sum_{i=1}^r\sum_{j=1}^s{n_{ij}\choose 2}\right\}
\end{equation}
which equals the proportion of unordered data pairs $\mathcal U=\{\{x_k,x_l\}\colon k,l=1,\dots,n,\, k\neq l\}$ that belong to the same cluster in one of the clusterings and to different clusters in the other clustering \citep[see][]{FS03}. This distance was called {\it symmetrical difference distance} in \cite{DG06}, and it is also considered in \citet{A15}, under the denomination {\it pairwise clustering loss}. In any case, it is not hard to check that ${\rm RD}(\mathscr C,\mathscr D)=\frac{n}{n-1}\widehat d_{\rm H}(\mathscr C,\mathscr D)$, so in fact there is little difference between these two empirical versions of $d_{\rm H}$.

Instead of measuring the dissimilarity between clusterings using a distance, clustering comparisons can be based on indices that quantify the agreement between them, with values close to 1 indicating greater similarity. In this sense, the Rand index \citep{R71} is defined as ${\rm RI}(\mathscr C,\mathscr D)=1-{\rm RD}(\mathscr C,\mathscr D)$. An important feature of the RI is that it also has a clear interpretation as the proportion of unordered data pairs that either belong to the same cluster or to different clusters in both clusterings. However, \citet{FM83} noted that, when comparing two clusterings with $r=s$, the range of possible values of the RI is quite narrow and its expected value $E({\rm RI})$ quickly approaches 1 as $r=s\to n$. This expectation is meant with respect to a random choice of the entries of the confusion matrix, while keeping its margins fixed, intended to reproduce a null scenario corresponding to independent clusterings. To amend this problem, \cite{HA85} proposed to correct the RI for chance, so that it yields an expected value of zero in such a null scenario, and introduced the adjusted Rand index ${\rm ARI}(\mathscr C,\mathscr D)=\{{\rm RI}(\mathscr C,\mathscr D)-E({\rm RI})\}/\{1-E({\rm RI})\}$. \cite{MC86} showed that, in addition, this correction also results in a much wider range of possible values for the ARI over the RI.

The most notable loss along this correction is the interpretation; for example, it is not easy to discern what an ARI value of $0.78$ means, or if an ARI of $0.82$ for two clusterings denotes a higher agreement between them than an ARI of $0.73$ for a different pair of clusterings, since the baseline $E({\rm RI})$ could be different. Precisely, in a series of papers (later collected in a single volume), \cite{GK79} emphasized the importance for association measures in cross classifications to have a clear operational interpretation. Besides, \cite{W83} raised some doubts with respect to the choice of the null scenario in the computation of $E({\rm RI})$ and, more recently, \cite{GA17} showed that the use of different null models for index adjustment can lead to disparate conclusions.

Despite these drawbacks, the ARI is one of the most popular and employed indicators for clustering comparison, in close competition with the MED. Hence, one of the main contributions of this paper is to provide a detailed inspection of both of them, via simple examples to help understanding their behaviour and their differences. Additional references concerning deep investigation of these criteria include \cite{W08}, \cite{SBH16} and \cite{SB18}, in the case of the ARI, and \cite{CDGH06}, \cite{CDH07} and \cite{D08} regarding the MED.

Here, since the MED is a distance and the ARI is an index, to facilitate their comparison the ARI will be previously transformed into a distance, which will be called the adjusted Rand distance and is defined as
$${\rm ARD}(\mathscr C,\mathscr D)=1-{\rm ARI}(\mathscr C,\mathscr D)={\rm RD}(\mathscr C,\mathscr D)/E({\rm RD}),$$
that is, as the Rand distance normalized by its expected value under the null model. This way, the ARD has unit expected value under the null model.

%
%

\section{Detailed comparison of the MED and the ARD}\label{sec:4}

In the following, several aspects of the MED and the ARD will be compared in detail. First, explicit computation of the two criteria is addressed. Then, the differences between the two are illustrated through several specific examples. Next, an exhaustive study of the simplest case of a $2\times 2$ confusion matrix is provided, with emphasis on exploring the most dissimilar situation between two clusterings. Finally, some of the lessons learned from the $2\times 2$ case are generalized for two clusterings of arbitrary size.

\subsection{Computation}\label{sec:comp}

One undeniable advantage of the ARD over the MED is its simpler definition, which readily translates into a much simpler computation.

Indeed, let us write $x_k\sim_{\mathscr C}x_l$ if the data points $x_k$ and $x_l$ belong to the same cluster in $\mathscr C$ (and $x_k\nsim_{\mathscr C}x_l$ otherwise), and consider the cardinalities of the sets of (unordered) data pairs that cover all the possibilities of belonging either to the same or to different clusters in $\mathscr C$ and $\mathscr D$, denoted
\begin{align*}
a&=\big|\{\{x_k,x_l\}\in\mathcal U\colon x_k\sim_{\mathscr C}x_l,\,x_k\sim_{\mathscr D}x_l\big\}\big|,\\
b&=\big|\{\{x_k,x_l\}\in\mathcal U\colon x_k\sim_{\mathscr C}x_l,\,x_k\nsim_{\mathscr D}x_l\big\}\big|,\\
c&=\big|\{\{x_k,x_l\}\in\mathcal U\colon x_k\nsim_{\mathscr C}x_l,\,x_k\sim_{\mathscr D}x_l\big\}\big|,\\
d&=\big|\{\{x_k,x_l\}\in\mathcal U\colon x_k\nsim_{\mathscr C}x_l,\,x_k\nsim_{\mathscr D}x_l\big\}\big|.
\end{align*}
Then, it is clear that ${\rm RD}(\mathscr C,\mathscr D)={n\choose 2}^{-1}(b+c)$. Moreover, \cite{St04} provided the very simple formula $E({\rm RI})={n \choose 2}^{-2}\{(a+b)(a+c)+(c+d)(b+d)\}$, which entails that $E({\rm RD})={n \choose 2}^{-2}\{(a+b)(b+d)+(a+c)(c+d)\}$, so that
\begin{equation}\label{eq:ARD}
{\rm ARD}(\mathscr C,\mathscr D)={n \choose 2}(b+c)/\{(a+b)(b+d)+(a+c)(c+d)\}.
\end{equation}
This is very easy to implement, taking into account that $a=\sum_{i=1}^r\sum_{j=1}^s{n_{ij} \choose 2}$, $b=\sum_{j=1}^s{n_{+j} \choose 2}-a$, $c=\sum_{i=1}^r{n_{i+} \choose 2}-a$ and $d={n\choose 2}-a-b-c$ can be immediately computed from the confusion matrix \cite[][Section 4.4.1]{JD88}.

In contrast, computation of the MED requires solving a discrete minimization problem over $\max\{r!,s!\}$ possible inputs, so its implementation is not that simple, which surely hinders its usage. To fully describe the problem, assume that $r\leq s$ and define $n_{i+}=n_{ij}=0$ for all $i=r+1,\dots,s$ (if any). Writing $m_{ij}=n_{i+}+n_{+j}-2n_{ij}$ for $i,j=1,\dots,s$, then computation of the MED involves finding $\min_{\sigma\in\mathcal P_s}\sum_{i=1}^sm_{i,\sigma(i)}$, where $\mathcal P_s$ denotes the set of all possible permutations of $s$ elements. Despite its apparent complexity, this is a form of the well-known assignment problem, and very efficient algorithms exist to find its solution \citep[see][]{BDM09}. Appendix A below offers a simple implementation using the popular R language \citep{R18}.

\subsection{Examples}

To help understanding what the ARD and the MED represent and how they are computed in practice it is useful to start with some simple real data examples.

The first example regards the famous iris data set \citep{A35}, including 4 measurements on $n=150$ flowers of three species of iris: {\it Iris setosa}, {\it versicolor} and {\it virginica}. If these data are clustered, e.g., using a normal mixture model \citep{FR02} with $G=3$ components, it results in the confusion matrix given in Table \ref{tab:iris}.

\begin{table}[ht]
\centering
\begin{tabular}{lrrr}\hline
&\multicolumn{3}{c}{Data-based labels}\\
  \cline{2-4}
 True labels& 1& 2 & 3 \\
  \hline\hline
\text{\it Setosa} & 50 & 0 & 0\\
\text{\it Versicolor} & 0 & 48 & 2\\
\text{\it Virginica}& 0 & 1 & 49\\
\hline\hline
\end{tabular}
\caption{Confusion matrix for normal mixture clustering against the true cluster labels for the iris data set.}
\label{tab:iris}
\end{table}

Thus, ${\rm MED}=(2+1)/150=0.02$ since only 3 data points would need to be re-labeled for the two partitions to coincide. On the other hand, there are $(2+1)\times(48+49)=291$ data pairs that belong to the same cluster in one of the partitions and to different clusters in the other, and that accounts for a proportion of ${\rm RD}=291/{150\choose 2}=0.026$ of the total number of possible data pairs. Finally, using Equation (\ref{eq:ARD}) the adjusted Rand distance for those two partitions is ${\rm ARD}=0.059$.

This is a very simple example because the two partitions have the same (small) number of clusters, and besides, they are quite similar. Nevertheless, it is helpful to perceive the differences between the MED, the RD and the ARD. Here, perhaps the MED is the easiest criterion to compute and to interpret, since it only involves counting misplaced {\it individual data points}. Obtaining the RD from the confusing matrix (by eye) is a bit more complex, since it implicates counting {\it data pairs}. And the corrected version ARD lacks the interpretability of the former two, but it still yields a very small number, indicating that the two partitions have a high degree of agreement.

Our second example concerns the DLBCL data set, introduced in \cite{Aal13}. It contains the records of the CD3, CD5 and CD19 antibodies on a set of $n=8183$ cells of a patient with Diffuse Large B-cell Lymphoma (DLBCL), along with the true cluster labels in five groups ($A$ to $E$) manually found by an expert. In \cite{Ch18}, this data set was analyzed using several component merging techniques for mixture model clustering, in particular through the so-called {\tt modclust} and {\tt entmerge} methods. The former suggested the existence of three clusters, while the latter correctly identified five clusters; both confusion matrices are given together in Table \ref{tab:dlbcl}.

\begin{table}[ht]
\centering
\begin{tabular}{lrrr||rrrrr}\hline
&\multicolumn{3}{c||}{modclust labels}&\multicolumn{5}{c}{entmerge labels}\\
  \cline{2-4}\cline{5-9}
 True labels& 1& 2 & 3 & 1 &2 & 3& 4& 5\\
  \hline\hline
$A$ & 47 & 197 & 7& 16  &  7 &   0 &  14 & 214\\
$B$ & 0 & 1408 & 153 & 0 & 146 & 929 & 417 &  69\\
$C$ &0 & 278 & 1216 & 0 & 1191 &  81 &  63 & 159\\
$D$ & 0 & 62 & 0 & 0  &  0  &  0 &   0 &  62\\
$E$ & 4813 & 2 & 0& 4809  &  0  &  0  &  1 &   5\\
\hline\hline
\end{tabular}
\caption{Confusion matrices for the clusterings obtained by {\tt modclust} and {\tt entmerge} for the DLBCL data set, as compared to the true cluster labels.}
\label{tab:dlbcl}
\end{table}

Regarding the confusion matrix for the {\tt modclust} labels, again it is not hard to compute the MED: the group matching leading to a higher degree of agreement would be $B$ with 2, $C$ with 3 and $E$ with 1, whereas the remaining $47+197+7+153+278+62+2=746$ data points would need to be re-labeled to make the two partitions coincide, thus yielding ${\rm MED}=746/8183=0.091$. Similarly, for the {\tt entmerge} labels it can be checked that ${\rm MED}=1040/8183=0.127$, so that the {\tt modclust} clustering is closer to the true expert labels regarding the MED, despite showing a smaller number of clusters. The reason is that, despite the {\tt entmerge} method returned the true number of clusters, its assignments to clusters 4 and 5 were so unfortunate (especially, the splitting of cluster $B$ into two significant groups in clusters 3 and 4) that a high number of re-labelings is needed to make this partition equal to the true one. In contrast, the ARD for these two confusion matrices can be computed to be ${\rm ARD}=0.112$ and ${\rm ARD}=0.097$ for the {\tt modclust} and {\tt entmerge} partitions, respectively. As noted before, this does not yield such an intelligible comparison regarding the relative closeness of the two data-based partitions to the true clustering, because the baseline $E({\rm RD})$ is different for the two contingency tables. Nevertheless, it must be noted that the unadjusted distances ${\rm RD}=0.055$ and ${\rm RD}=0.047$ (respectively) also suggest that, in terms of data-pair disagreements, the {\tt entmerge} clustering seems to be slightly closer to the expert partition than the {\tt modclust} one.

The two previous examples illustrate the common scenario in real data analysis, where data-based partitions are not too dissimilar from the true clustering. To finish this section, a synthetic example concerning quite distant partitions is examined. The confusion matrix shown in Table \ref{tab:steinley} corresponds to the two assignments of $n=13$ objects into $r=s=5$ clusters in Table 2 in \citet{St03}.

\begin{table}[t!h]
\centering
\begin{tabular}{lrrrrr}\hline
&\multicolumn{5}{c}{ Clustering $\mathscr D$}\\
  \cline{2-6}
 Clustering $\mathscr C$& $D_1$& $D_2$ & $D_3$ & $D_4$ & $D_5$\\
  \hline\hline
$C_1$ & 1 & 0 & 1 & 1 & 0\\
$C_2$ & 0 & 1 & 0 & 0 & 1\\
$C_3$ & 1 & 0 & 1 & 0 & 1\\
$C_4$ & 0 & 1 & 0 & 1 & 0\\
$C_5$ & 1 & 0 & 1 & 0 & 1\\
\hline\hline
\end{tabular}
\caption{Confusion matrix for Steinley's example.}
\label{tab:steinley}
\end{table}

To appreciate how distant these two clusterings are, it is worth noting that ${\rm ARD}=1.164$, greater than 1, meaning that the disagreement between the two is higher than the average that would be obtained if the labels were randomly assigned (following the null model). The number of data pairs that are in the same group in one clustering and in different groups in the other can be computed to be 22, out of the total of ${13\choose 2}=78$ possible data pairs, which leads to ${\rm RD}=22/78=0.282$. And, by considering any permutation of the columns of the confusion matrix that preserves all its diagonal entries as 1, adding up the off-diagonal figures leads to ${\rm MED}=8/13=0.615$.

This example further illustrates how counting ``discordant" data pairs seems to be less intuitive than counting ``discordant'' individual data points. But also, it shows that the permutation for which the MED is attained may not be unique: for instance, rearranging the columns of the confusion matrix according to the permutation $(45123)$ yields the same MED value, as already noted in \citet{St03,St04}. In any case, it is easy to check that the values of the RD and the ARD also remain the same under that permutation. Such a phenomenon is expected to occur for the comparison of very dissimilar clusterings; for example, in the extreme case where the confusion matrix ${\mathbf N}$ has all its entries equal to 1 (representing independent label assignments), then any permutation of its column leads to the same MED, RD and ARD values.


\subsection{Two clusters in each clustering}\label{sec:rs2}

In order to gain a deeper understanding of the behaviour of the MED and the ARD the next step is to analyze in detail the simplest scenarios. Arguably, the simplest comparison between two clusterings arises when either $r=1$ or $s=1$, but that could be considered a degenerate case, since in fact one of the partitions would show no clusters. So the next simplest case is $r=s=2$; we will focus our attention on this case first, and then we will generalize some of our findings to the case of arbitrary $r$ and $s$.

Independently of the criterion employed to compare clusterings, any researcher would probably agree that having a diagonal confusion matrix is synonymous with a perfect agreement between the two partitions. But that is also the case if the confusion matrix is anti-diagonal, which means, for $r=s=2$, that
$${\mathbf N}=\begin{pmatrix}0&n_{12}\\n_{21}&0\end{pmatrix}.$$
This clearly illustrates a key difference between classification and clustering: since classification is a supervised learning problem, the training data are already equipped with precise-meaning labels, and hence an anti-diagonal confusion matrix must be interpreted as the result of a totally wrong classification; in contrast, a clustering algorithm labels the groups as it finds them and, hence, the coding designation is not important (group 1 might as well have been called group 2, and viceversa) so that an anti-diagonal confusion matrix also represents perfect agreement, since the discovered groups are exactly the same, only differing in their (arbitrary) denomination. Mathematically, this means that distances between clusterings must be invariant with respect to permutations of the cluster labels \citep{M12}.

\begin{figure}\centering
\includegraphics[width=.5\textwidth]{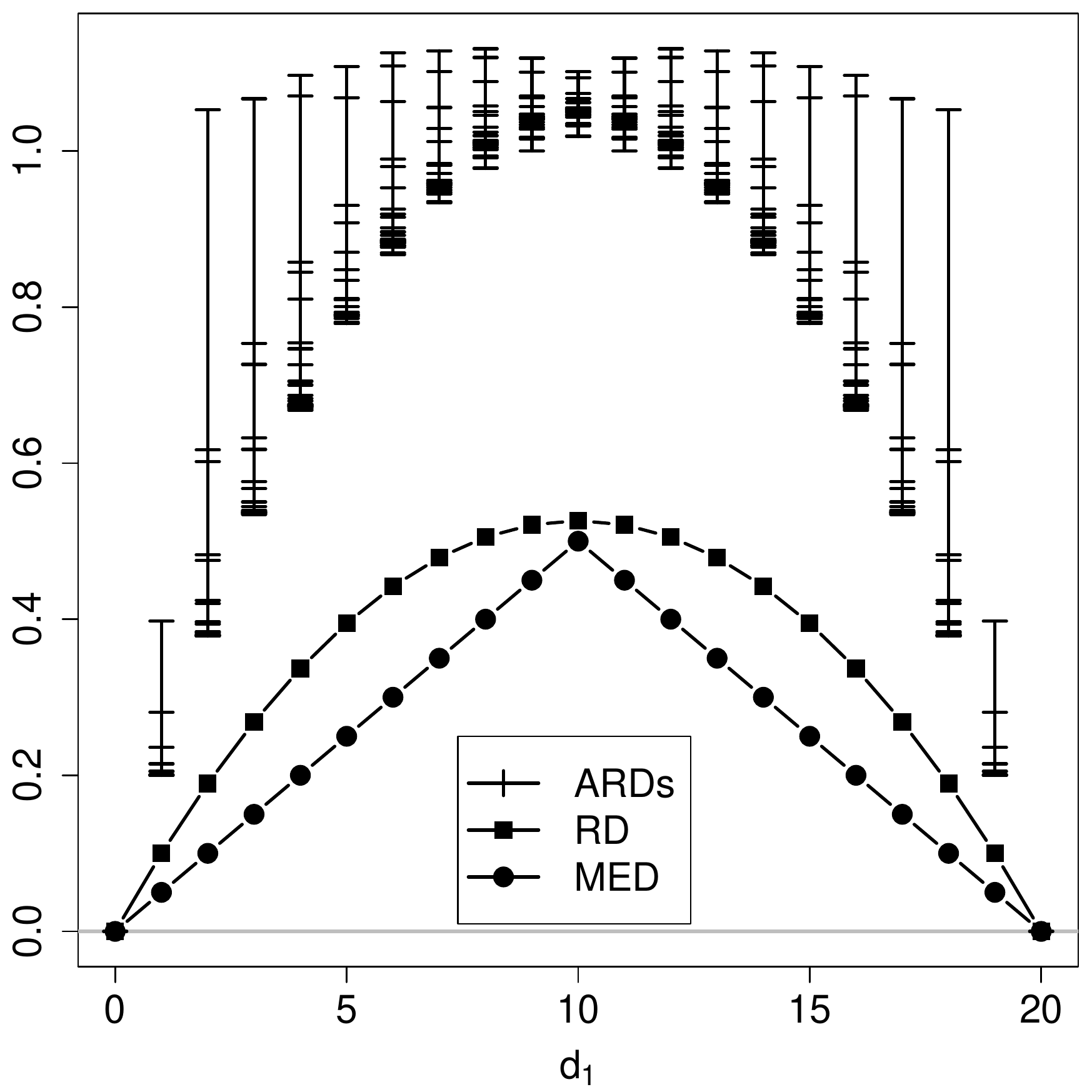}
\caption{MED (solid circles), RD (solid squares) and possible ARD values (ticks) as a function of $d_1$ for $n=20$.}
\label{fig:MED_RD}
\end{figure}

It is precisely the way to measure deviations from the diagonal or anti-diagonal situation what gives rise to the different distances between clusterings. For the case $r=s=2$, let us consider the $2\times 2$ confusion matrix ${\mathbf N}=(n_{ij})$, and denote $d_1=n_{11}+n_{22}$ and $d_2=n_{12}+n_{21}$ the total sum of its diagonal and anti-diagonal entries, respectively. In this case, it is not hard to show that the MED and the RD can be simply expressed as
\begin{align*}
{\rm MED}&=n^{-1}\min\{d_1,d_2\}\\
{\rm RD}&={n\choose 2}^{-1}d_1d_2.
\end{align*}
To graphically appreciate the differences between the MED and the RD, and noting that $d_2=n-d_1$, Figure \ref{fig:MED_RD} shows the possible values of these criteria for $n=20$, as a function of $d_1$. The linear and quadratic appearances of the MED and the RD, respectively, are explained by the fact that they can be equivalently expressed as ${\rm MED}=1/2-|d_1/n-1/2|$ and ${\rm RD}={n\choose 2}^{-1}d_1(n-d_1)$.

On the other hand, it is not possible to express the ARD as a function of $d_1$ and $d_2$ only. For a given value of $d_1$, there exist configurations of the confusion matrix that result in different ARD values. Figure \ref{fig:MED_RD} also shows all these possible ARD values for each given $d_1$ (marked with a tick over the whole possible range, that is indicated with a vertical line). This reveals a somehow erratic behaviour of the ARD in some cases, and inspecting such cases more closely allows to clarify how the ARD works. For instance, for $n=20$ consider the confusion matrices
$${\mathbf N}_1=\begin{pmatrix}16&2\\2&0\end{pmatrix},\quad{\mathbf N}_2=\begin{pmatrix}11&0\\4&5\end{pmatrix}.$$
The two matrices have $d_1=16$, so that ${\rm MED}=0.2$ and ${\rm RD}=0.337$ for both ${\mathbf N}_1$ and ${\mathbf N}_2$. However, ${\rm ARD}=1.097$ for ${\mathbf N_1}$ whereas ${\rm ARD}=0.668$ for ${\mathbf N}_2$. In the first configuration, in both clusterings there is a big cluster with 18 elements and a relatively small one with only 2 elements; both clusterings agree on most of the elements in the big cluster, but show no agreement at all regarding the small cluster, since none of the data points has been simultaneously assigned to the small cluster in both clusterings.  In the second configuration, the first clustering presents two quite balanced clusters, say $\mathscr C=\{C_1,C_2\}$, of sizes 11 and 9 (respectively), while the second clustering has clusters of sizes $15$ and $5$, which can be obtained from $\mathscr C$ by transferring 4 elements from $C_2$ to $C_1$. The ARD seems to penalize the first configuration much more severely than the second one.

\subsubsection{Worst-case scenario}

The previous formulas for the MED and the RD in terms of $d_1$ and $d_2$ are also useful to analyze the worst-case scenario; i.e., the situation in which two given clusterings are as dissimilar as possible. If $n$ is even, then the maximum possible MED is $1/2$ and it is attained for $d_1=d_2=n/2$. Thus, it is worth remarking that even for the two most dissimilar possible clusterings the MED is not going to be higher than $0.5$ for the case of $r=s=2$. This could make a case against the use of the MED, since one would expect this distance to attain a maximum of 1 when comparing the most dissimilar clusterings. However, a moment of reflection reveals that this maximum of $0.5$ makes perfect sense in the context of clustering comparison, due to the aforementioned feature that any cluster label permutation should not affect distances between clusterings: having a proportion of label disagreements greater than $0.5$ would mean than exchanging the label denominations we would get a proportion smaller than $0.5$.

Nevertheless, it is helpful to keep the value of the maximum possible distance in mind at the time of judging how far two clusterings are: a MED of $0.4$ always has the same interpretation, but in relative terms it represents a worse result if the maximum possible MED is $0.5$ than if it is $0.95$. Hence, this suggests the introduction of a normalized MED, defined as ${\rm NMED}={\rm MED}/\max{\rm MED}$, to record how large the MED is with respect to its maximum possible value (given fixed values of $r$, $s$ and $n$). This should not replace the unnormalized MED, since they offer different information, but they should be given together. In the previous example, having ${\rm MED=0.4}$, ${\rm NMED}=0.8$ versus ${\rm MED=0.4}$, ${\rm NMED}=0.42$ indicates that the former situation is closer to the case of totally dissimilar clusterings than the latter one. Notice that this is a very different adjustment from the usual one, since it is not based on the expected value of the index under some null model; indeed, it does not rely on any choice of a null model.

The difficulty of such a normalization is that it is necessary to analyze which is the worst-case scenario for each index. Continuing with the $2\times 2$ table, it is not hard to check that $\max{\rm MED}=(n-1)/(2n)$ if $n$ is odd, which is attained for both $d_1=(n-1)/2$ and $d_1=(n+1)/2$. Therefore, ${\rm NMED}=2n^{-1}\min\{d_1,d_2\}$ for even $n$ and ${\rm NMED}=2(n-1)^{-1}\min\{d_1,d_2\}$ for odd $n$. 

Regarding the RD, its maximum is attained at the same value of $d_1$ as for the MED, resulting in $\max{\rm RD}=n/\{2(n-1)\}$ for even $n$ and $\max{\rm RD}=(n+1)/(2n)$ for odd $n$, so that it approaches $1/2$ as $n$ increases. Hence, the normalized RD, defined as ${\rm NRD}={\rm RD}/\max{\rm RD}$, can be explicitly written as ${\rm NRD}=4n^{-2}d_1d_2$ for even $n$ and ${\rm NRD}=4\{(n-1)(n+1)\}^{-1}d_1d_2$ for odd $n$. 

For the ARD, it would have been expected that its maximum were attained amongst the possible configurations with $d_1=n/2$ for even $n$ (or $d_1=(n-1)/2$ for odd $n$), but Figure \ref{fig:MED_RD} shows that this does not happen, in general. For instance, for $n=20$ the maximum ARD is attained for a configuration with $d_1=12$; more precisely, for ${\mathbf N}=\left(\begin{smallmatrix}12&4\\4&0\end{smallmatrix}\right)$, which gives $\max{\rm ARD}=95/84\simeq1.131$. It is somehow counterintuitive that the maximum value of the ARD is not attained for ${\mathbf N}=\left(\begin{smallmatrix}5&5\\5&5\end{smallmatrix}\right)$, which represents the situation where the labels of the first clustering are perfectly independent from the labels in the second clustering.

In fact, it would be interesting to study which are the possible maximum and minimum values of the ARD for a given $d_1$. Since $d_1$ and $d_2=n-d_1$ are fixed, the numerator in the definition of the ARD is constant, so this problem is equivalent to finding the minimum and maximum values of $E({\rm RD})$ for a given $d_1$. It appears (although it was not possible to find a simple proof) that for a given $d_1\geq d_2$, the maximum ARD is attained for
\begin{equation}\label{eq:maxARD}
{\mathbf N}=\begin{pmatrix}d_1&d_2/2\\d_2/2&0\end{pmatrix}, \quad\text{or}\quad{\mathbf N}=\begin{pmatrix}d_1&(d_2-1)/2\\(d_2+1)/2&0\end{pmatrix},
\end{equation}
provided $d_2\geq2$ is even or odd, respectively, and that for $d_1=n-1,\,d_2=1$ the confusion matrix configuration that maximizes the ARD is $\left(\begin{smallmatrix}n-2&0\\1&1\end{smallmatrix}\right)$. Using the form for even $d_2$, the resulting maximum ARD for a given $d_1$ can be expressed as
$$\alpha_n(d_1)=\frac{4n(n-1)d_1}{(d_1+n)\{d_1^2+n(n-2)\}}$$
for $d_1\geq d_2\geq 2$. Maximizing $\alpha_n(d_1)$ with respect to $d_1$ yields $\max{\rm ARD}$, but it is not clear how to obtain an explicit expression for such a maximum. In any case, note that it is not necessary to normalize the ARD, since this distance already includes a kind of normalization (although by the expected value of the RD, not by its maximum).

\subsubsection{Close clusterings}

Similarly, this in-depth inspection of the $2\times2$ case is also beneficial to understand how these measures of dissimilarity between two clusterings evolve when such clusterings are very close. All these distances obviously return a zero value if $n_{12}=n_{21}=0$, but the question that will be addressed here is how these distances behave as $n_{12}\to0$ and $n_{21}\to0$ before reaching their null limit. More precisely, the goal is to provide a linear approximation of the MED and the RD for small values of $n_{12}$ and $n_{21}$.

Such an approximation is very easy to find for the MED, since as both $n_{12},n_{21}\to0$ it is clear that $\min\{d_1,d_2\}=n_{12}+n_{21}$, so that ${\rm MED}=(n_{12}+n_{21})/n$ for small values of $n_{12}$ and $n_{21}$ (this is an equality rather than an approximation). On the other hand, it is possible to write ${\rm RD}={n\choose 2}^{-1}\big\{n(n_{12}+n_{21})-(n_{12}+n_{21})^2\big\}$, so that a Taylor expansion gives ${\rm RD}\simeq2(n_{12}+n_{21})/(n-1)$ as $n_{12},n_{21}\to0$. This means that, for small values of $n_{12}$ and $n_{21}$, the RD will be roughly twice the MED.

For instance, for ${\mathbf N}=\left(\begin{smallmatrix}55&6\\4&35\end{smallmatrix}\right)$ we have ${\rm MED}=0.1$ and ${\rm RD}=0.182$, while the approximation formula for the RD reads $2\cdot(6+4)/(100-1)=0.202$.

\subsection{Arbitrary number of clusters}\label{sec:arbitrary}

The case $r=s=2$ is surely the easiest one to analyze in detail, and its analysis results in a deeper understanding of how the MED, the RD and the ARD behave. Here, such an analysis is extended for the comparison of two clusterings with an arbitrary number of clusters.

One of the findings for $r=s=2$ is that the MED attains its maximum when the clustering labels are perfectly independent. In general, this refers to the situation where $n$ is a multiple of $rs$ and the $(r\times s)$-confusion matrix ${\mathbf N}$ has all its entries equal to $n/(rs)$. In that case, note that $\sum_{i=1}^rn_{i,\sigma(i)}=n/s$ for any $\sigma\in\mathcal P_s$ and, therefore, ${\rm MED}=1-1/s$. But, assuming $r\leq s\leq n$, \citet[][Lemma 1]{CDGH06} showed that an upper bound for the MED is $1-\lceil n/s\rceil/n$ (with $\lceil\cdot \rceil$ standing for the ceiling function), which generalizes the bounds obtained for odd or even $n$ in the previous section for $s=2$. Hence, once again the maximum MED is attained for the case of perfectly independent clustering labels. Thus, the corresponding normalized MED is defined in general as ${\rm NMED}=\frac n{n-\lceil n/q\rceil}{\rm MED}$, where $q=\max\{r,s\}$. The effect of this normalization is noticeable for small values of $q$, but becomes negligible as $q$ increases.

In contrast, the story for the RD with arbitrary $r\leq s$ is different from the case $r=s=2$. An exhaustive enumeration of all the possible confusion matrix configurations for small values of $r$, $s$ and $n$ suggests that, given $r\leq s$ and $n\geq 2(r-1)+s$, the maximum value of the RD is always attained for a matrix of the form
\begin{equation}\label{eq:argmaxRD}
{\mathbf N}=\left(\begin{array}{cccc}
q_1&q_2&\cdots&q_s\\
1&0&\cdots&0\\
\vdots&\vdots&\ddots&\vdots\\
1&0&\cdots&0\\
\end{array}\right),
\end{equation}
with $q_1,\dots,q_s\in\mathbb N$ and $q_1\geq\dots\geq q_s$ (it remains open to find a formal proof of this fact). This does not mean that the maximizing matrix is necessarily unique; in fact, as noted before, for $r=s=2$ and $n=20$ the maximum RD is attained for any confusion matrix with $d_1=10$, for instance for ${\mathbf N}=\left(\begin{smallmatrix}10&9\\1&0\end{smallmatrix}\right)$ which has the form (\ref{eq:argmaxRD}), and also for ${\mathbf N}=\left(\begin{smallmatrix}5&5\\5&5\end{smallmatrix}\right)$, which represents the perfectly independent situation.
In addition, assuming that the conjectured form of the maximizer is correct, it is shown in Appendix B that $\max{\rm RD}$ is attained by taking
$q_1=k+r-1$, the next $\ell\geq0$ coordinates $q_2=\cdots=q_{\ell+1}=k+1$ and the remaining $s-\ell-1\geq0$ coordinates $q_{\ell+2}=\cdots=q_s=k$, where $k=\lfloor \{n-2(r-1)\}/s\rfloor\in\mathbb N$ (with $\lfloor\cdot\rfloor$ standing for the floor function) and $\ell=n-2(r-1)-ks\in\{0,1,\dots,s-1\}$; that is, $k$ and $\ell$ are the quotient and the remainder of the (Euclidean) division of $n-2(r-1)$ by $s$, respectively. With such a choice, it follows that
\begin{equation}\label{eq:maxRD}
n(n-1)\max{\rm RD}=(n-r+1)^2+(r-1)(2r-3)-sk^2-l(2k+1).
\end{equation}
This allows to explicitly define the normalization ${\rm NRD}={\rm RD}/\max{\rm RD}$.

\begin{figure}\centering
\includegraphics[width=.5\textwidth]{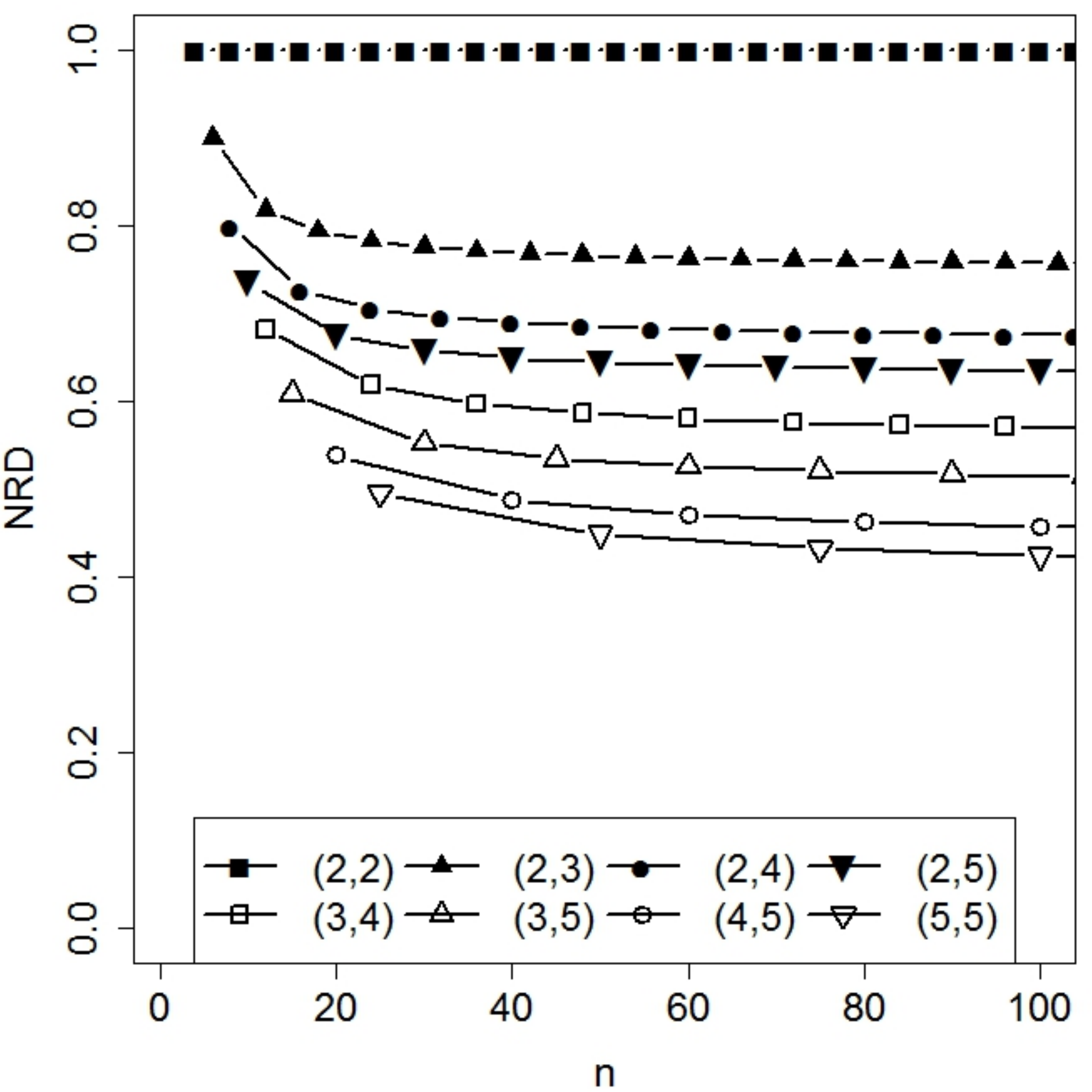}
\caption{Normalized RD for perfectly independent clusterings with $r$ and $s$ clusters, as a function of the sample size $n$, for several combinations of $(r,s)$ as indicated in the legend.}
\label{fig:NRDi}
\end{figure}

Further, when $n$ is a multiple of $rs$ and the two clusterings are perfectly independent it is easy to check that $n(n-1){\rm RD}=n^2(r+s-2)/(rs)$. Since $\max{\rm RD}\sim1-1/s$ as $n\to\infty$, it follows that the maximum ${\rm RD}$ is not attained for perfectly independent clusterings for big enough $n$ if $s>2$. Moreover, in practice this seems to be the case for all $n$, as shown in Figure \ref{fig:NRDi}. This figure represents the normalized RD for the case of two perfectly independent clusterings for several combinations of $r$ and $s$. Only for the pair $(r,s)=(2,2)$ the RD for perfectly independent clusterings matches its maximum possible value. For any other combination, having two totally unrelated clusterings does not yield the maximum possible RD; indeed, this phenomenon becomes more and more severe as $r+s$ increases and, for instance, for $(r,s)=(5,5)$ and $n=100$ the confusion matrix with all its entries equal to 4 results in a RD that is only $42\%$ of the maximum achievable RD, attained for a matrix of the form (\ref{eq:argmaxRD}) with $q_1=22$, $q_2=q_3=19$ and $q_4=q_5=18$.

Finally, it is worth noting that $(r+s-2)/(rs)$ decreases as $r$ and/or $s$ increases, so that the RD for the case of perfectly independent clusterings becomes quite small when both $r$ and $s$ are large and, hence, the RD does not seem useful to detect this important instance of unrelated clusterings. \citet[][p. 555]{FM83} already noted this phenomenon, upon inspecting the expected value and variance of the Rand index under the null model.


For the ARD, it was not possible to provide an explicit formula for its maximum for $r=s=2$, and the problem is of course more intricate for arbitrary $r$ and $s$. Nevertheless, it seems clear that the maximum ARD is not attained for the case of independent clustering labels, in general. Instead, the inspection of all possible confusion matrix configurations for small values of $r$, $s$ and sufficiently large $n$ seems to suggest that the maximum value of the ARD is always attained for a matrix of the form
$${\mathbf N}=\begin{pmatrix}p_1&q_2&\cdots&q_s\\
p_2&0&\cdots&0\\
\vdots&\vdots&\ddots&\vdots\\p_r&0&\cdots&0\end{pmatrix},$$
with $p_1,\dots,p_r,q_2,\dots,q_s\in\mathbb N$ and $p_1\geq p_2\geq\cdots\geq p_r$, $p_1\geq q_2\geq\cdots q_s$ (furthermore, with $(p_2,\dots,p_r)=(q_2,\dots,q_s)$ if $r=s$).
Indeed, confusion matrices with ARD larger than the value corresponding to the perfectly independent case can be constructed by following the guidelines described above for $r=s=2$. For instance, if $n=24$, $r=2$, $s=3$, then the confusion matrices
$${\mathbf N}_1=\begin{pmatrix}4&4&4\\4&4&4\end{pmatrix}\quad\text{and}\quad{\mathbf N}_2=\begin{pmatrix}15&4&1\\4&0&0\end{pmatrix}$$
lead to ARDs of $1.062$ and $1.143$, respectively, and for $n=27$, $r=3$, $s=3$, the confusion matrices
$${\mathbf N}_3=\begin{pmatrix}3&3&3\\3&3&3\\3&3&3\end{pmatrix}\quad\text{and}\quad{\mathbf N}_4=\begin{pmatrix}15&5&1\\5&0&0\\1&0&0\end{pmatrix}$$
yield ARDs of $1.083$ and $1.157$, respectively. Moreover, when $n$ is a multiple of $rs$ and the two clusterings are perfectly independent, it is easy to show that ${\rm ARD}=(n-1)/\{n-(2rs-r-s)/(r+s-2)\}$, which approaches 1 (from above) as $n$ increases.

\section{Numerical experiments}\label{sec:5}

In this section, the distributions of the MED, the RD, the ARD and the normalized versions NMED and NRD will be compared in different simulated scenarios.

As noted in \cite{VMal18}, benchmarking studies for cluster analysis do not abound. Nevertheless, the task of comparing different external criteria via simulation was addressed in the seminal paper by \cite{MC86} and also more recently in \cite{St04}, \cite{DG06} or \cite{SB18}.

Broadly speaking, these studies handle two possible scenarios. The first one explores the performance of the criteria in the null case, that is, when the agreement between the compared clusterings is only due to chance. And the second framework concerns how the criteria of interest behave as the two compared clusterings drift apart, starting from perfect similarity. Both scenarios are considered separately in the next sections.

\subsection{The null case}

As noted above, the null case scenario covers the situation where the clustering agreements are solely due to chance. However, as remarked in \cite{GA17}, different choices for the model for random clusterings can be made, and a careful model selection is needed to provide a baseline that is neither based on a model that ``is not random enough'' nor on a model that is ``too random''.

These authors considered three models for random clusterings, with increasing level of randomness, starting with the permutation model (where the number of clusters and their sizes is fixed), followed by the model where only the number of clusters is fixed, and finally the model encompassing all possible clusterings, with arbitrary number of clusters and cluster sizes. As a compromise for intermediate randomness level, in this section the null case refers to the situation where random labels are drawn uniformly after fixing the number of clusters.

\begin{figure}[t]\centering
\includegraphics[width=.8\textwidth]{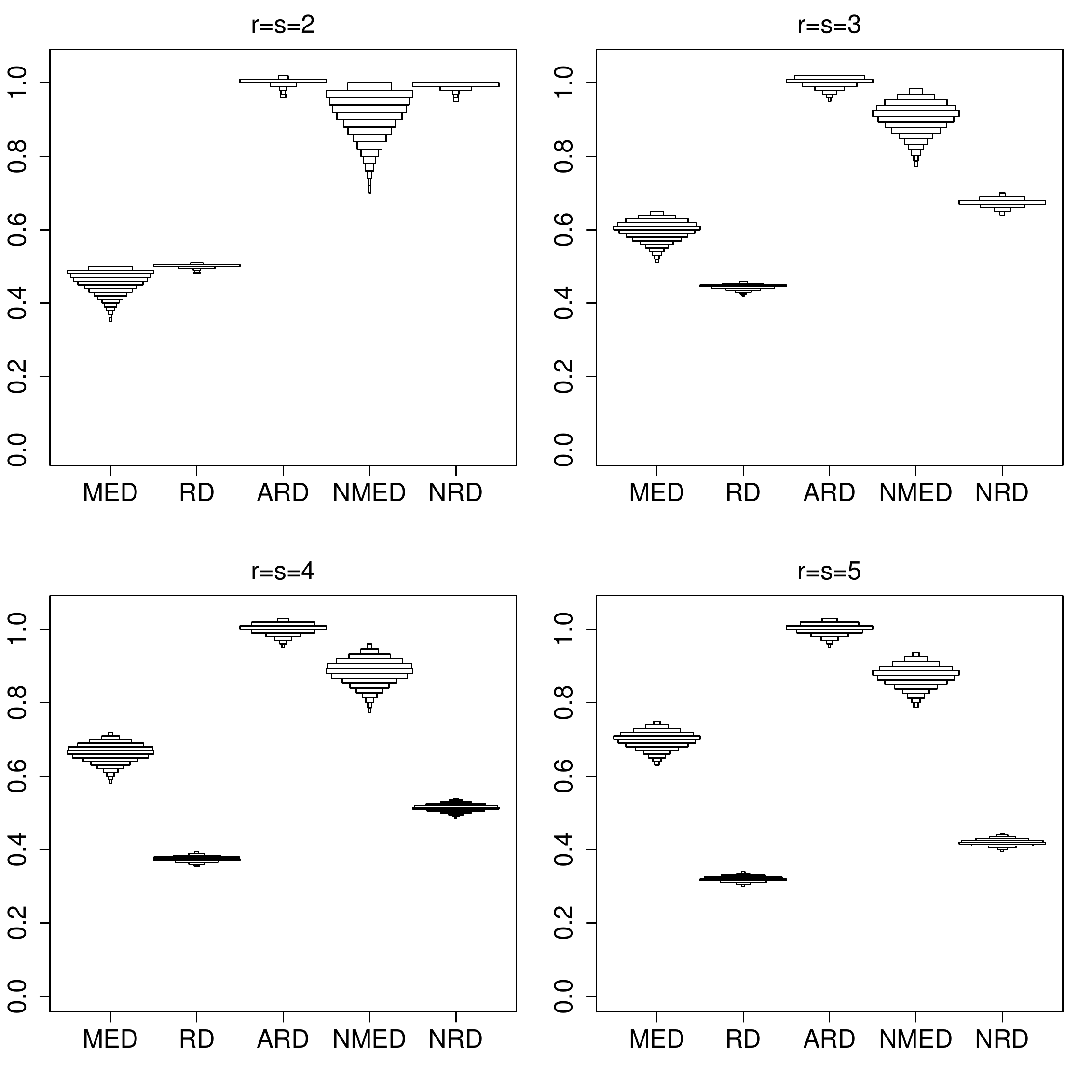}
\caption{Distribution of the criteria in the null case for sample size $n=100$.}
\label{fig:null100}
\end{figure}

\begin{figure}[t]\centering
\includegraphics[width=.8\textwidth]{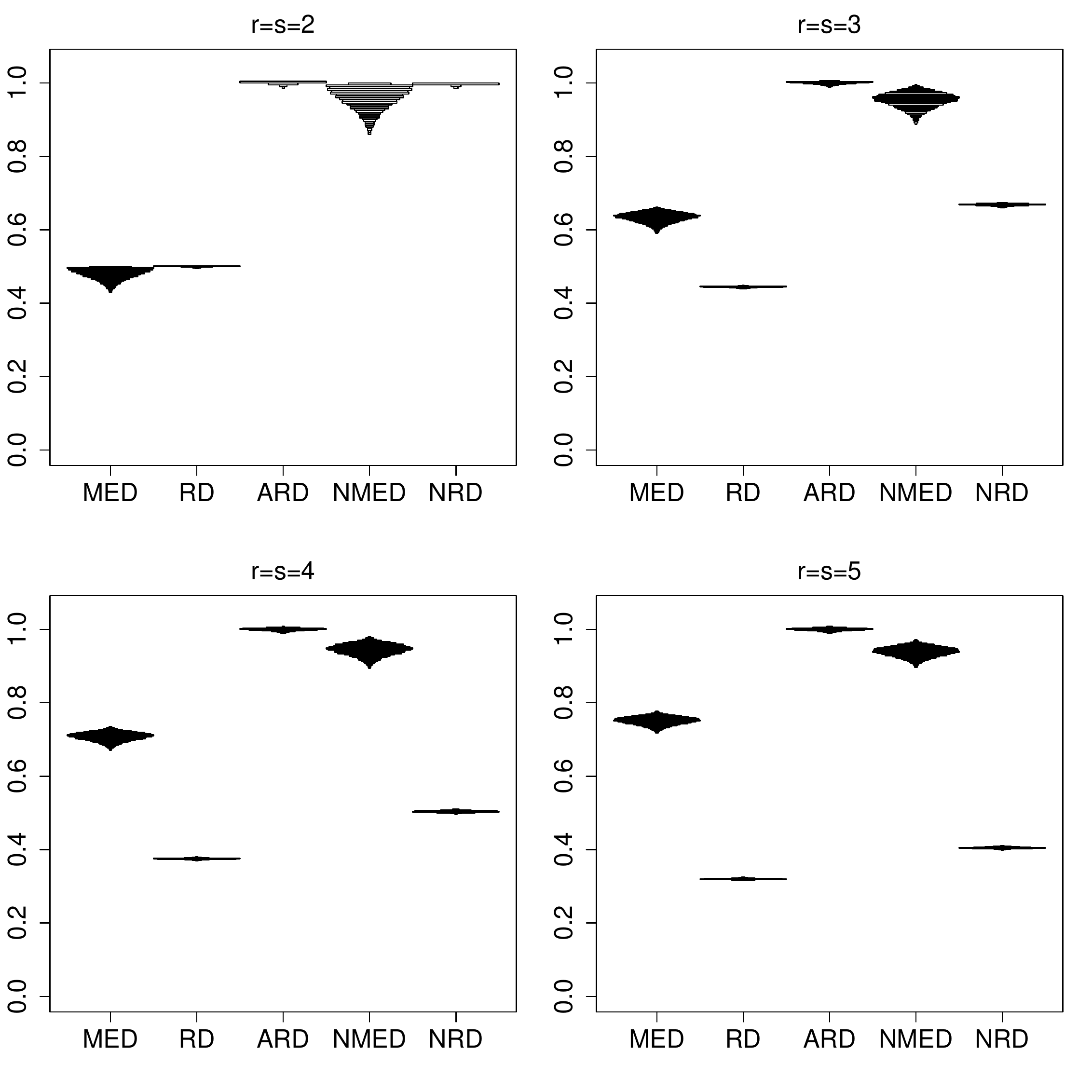}
\caption{Distribution of the criteria in the null case for sample size $n=400$.}
\label{fig:null400}
\end{figure}

Hence, the distribution of the considered criteria in the null case is explored by computing their values on a big enough number $B$ of random clustering pairs of $n$ objects, obtained by independently drawing two uniform samples of size $n$ from $\{1,\dots,r\}$ and $\{1,\dots, s\}$, respectively. The number of synthetic replicates was set to $B=10000$, in order to obtain a precise approximation of the distributions; the number of clusters was considered equal ($r=s$), in common with some of the aforementioned previous studies, and ranging in $\{2,3,4,5\}$; and samples sizes $n=100$ and $n=400$ were used to investigate the effect of an increasing number of data points. The distributions of the studied criteria are depicted in Figures \ref{fig:null100} and \ref{fig:null400} for $n=100$ and $n=400$, respectively, by means of side-by-side vertical histograms (with the bars mirrored with respect to the vertical axis), whose bars have been rescaled so that each of them has maximum bar length equal to 1, to aid visualization.


Despite being corrected for chance according to the permutation model (which is not exactly the null model in this study), the distribution of the ARD seems to be centered at 1 in all cases, so this type of adjustment makes it possible to compare its behaviour along the different configurations. Its variability is the second lowest among the compared criteria, it seems not to change with the number of clusters but quickly decreases with the sample size, as also noted in \cite{SBH16}. This suggests that the ARD may give rise to a powerful tool for detecting clustering independence.

The RD is the least variable criterion out of those considered here. This is not surprising in view of the previous Figure \ref{fig:MED_RD}, since its quadratic nature entails a least pronounced descent around its null-case value than the MED, for instance. Besides, as remarked in the previous section, under this null scenario the RD only achieves its maximum value for the case $r=s=2$, which yields a tightly concentrated distribution of the NRD with a maximum of 1 in that case. However, as previously shown in Figure \ref{fig:NRDi}, the maximum possible value of the RD becomes quite bigger than its value for independent clusterings as the number of clusters increases, and this explains why even the distributions of the normalized RD are far from 1. In other words, confusion matrices corresponding to randomly generated clusterings are usually far from something like (\ref{eq:argmaxRD}). Probably, it might be possible to obtain NRD distributions much closer to 1 if the random clusters were generated to produce confusion matrices only slightly deviated from (\ref{eq:argmaxRD}), but that does not seem to be an appropriate null model.

The MED is notably more variable than the ARD and the RD, with standard deviations about 1.6--2.1 times greater than those of the ARD, and 3.5--4.2 times greater than those of the RD, for $n=100$ (3.1--4.2 and 6.7--8.5, respectively, for $n=400$). Its variability, though, appears to decrease slightly as the number of clusters grow. Its approximated distribution shows an upper bound that agrees with the results in the previous section (for $n=100$, e.g., a maximum value of 0.5, 0.66, 0.75 and 0.8 for $r=s=2,3,4,5$, respectively), yielding location features that naturally change with the number of clusters, and hence making inappropriate to aggregate its results across the different simulated configurations. This upper bound also implies that NMED certainly attains a maximum value of 1 for this null scenario of random clusterings. However, it must be pointed out that the probability of attaining such a maximum value seems to decrease with the number of clusters.

Indeed, in some cases it is possible even to give an exact expression for such a probability. For $r=s=2$ and even $n$, for instance, it corresponds to $P(d_1=n/2)$, where $d_1$ is the sum of the two diagonal terms in the confusion matrix. In the null scenario, $d_1$ is a random variable following a binomial distribution, with $n$ as the number of trials and probability of success $p=1/2$ (the probability that two uniform and independent choices from $\{1,2\}$ are the same). Hence, $P(d_1=n/2)={n\choose n/2}\big/2^n$. More generally, here the random variable $n\cdot{\rm MED}$ follows a folded binomial distribution \citep{G70}.

\subsection{Diverging clusterings}\label{sec:diverging}

The second simulation scenario concerns studying the evolution of the compared criteria as two clusterings move away from each other, starting from a situation of perfect agreement.

Interestingly, in most of the existing simulation studies \citep[see, for instance,][]{St04,DG06}, the process of ``moving away from each other'' is quantified by measuring the proportion of data points that are differently clustered from the initial stage of perfect agreement. \cite{St04} called this proportion the ``degree of overlap'', and more recently \citet[][Section 3.2.2]{SB18} referred to this measure of deviation from the perfect agreement as the misclassification rate. So, overall, this simulation scenario concerns inspecting how the other clustering distances evolve as compared to the MED (see Figure 3 in \citealp{St04}, or Figure 1 in \citealp{DG06}).

Aside, it should be noted that what \cite{St04} and \cite{SB18} called degree of overlap is not exactly the same as the MED. The simulation setup in these references concerns a diagonal confusion matrix as the starting point (hence, a situation of perfect agreement) that is progressively perturbed by randomly taking a proportion of objects from the diagonal and placing them in off-diagonal cells. This proportion of off-diagonal objects is what is called degree of overlap (DO), and in the aforementioned studies it is allowed to vary in $0.05, 0.10, \dots, 0.95$. 
However, this is not the same as the misclassification rate: while the DO and the MED usually coincide for low DO values, when the DO is too high it may occur that the resulting clusterings became indeed closer with respect to the MED, instead of further away. For example, consider the confusion matrices
$${\mathbf N}_1=\begin{pmatrix}8&0&0\\0&6&0\\0&0&6\end{pmatrix},\quad {\mathbf N}_2=\begin{pmatrix}3&2&3\\2&2&2\\2&2&2\end{pmatrix},\quad{\mathbf N}_3=\begin{pmatrix}1&2&5\\3&1&2\\2&4&0\end{pmatrix}.$$
For all of them, $n=20$. Matrix ${\mathbf N}_2$ stems from ${\mathbf N}_1$ after removing a total of 13 objects from the diagonal, so that ${\rm DO}=13/20=0.65$ for ${\mathbf N}_2$; it can be checked that ${\rm MED}=0.65$ for ${\mathbf N}_2$ as well. Five additional objects are removed from the diagonal when going from ${\mathbf N}_2$ to ${\mathbf N}_3$, representing a total ${\rm DO}=18/20=0.9$ with respect to ${\mathbf N}_1$, but ${\rm MED}=0.4$ for ${\mathbf N}_3$, lower than for ${\mathbf N}_2$. Of course, this is due to the fact that $\max{\rm MED}=0.65$ for $n=20$ and $r=s=3$, so that it does not seem appropriate to consider DO values greater than $0.65$ in this case.


In \cite{DG06}, several agreement indices were compared as a function of an increasing MED. A given starting partition is recursively perturbed by randomly selecting one element and a new class label for it. This procedure aims at randomly and equiprobably generating partitions at precise MED of the given one. However, there the number of clusters is not fixed and, hence, their study comprises a higher degree of uncertainty.

Nevertheless, for the goal of inspecting the evolution of the different distances as a function of the MED, the most exhaustive procedure is surely that based on computing the involved measures for all the possible confusion matrix configurations, that is, for all the matrices in $\mathcal N(r,s,n)=\{{\mathbf N}\in\mathcal M_{r\times s}\colon  n_{i+}>0$ for all $i$, $n_{+j}>0$ for all $j$, and $\sum_{i,j}n_{ij}=n\}$. Indeed, this is precisely what Figure \ref{fig:MED_RD} represents for $r=s=2$ and $n=20$. But this could be accomplished in that case because the cardinality $|\mathcal N(2,2,20)|=1691$ was reasonably small.

\begin{figure}[t]\centering
\includegraphics[width=.7\textwidth]{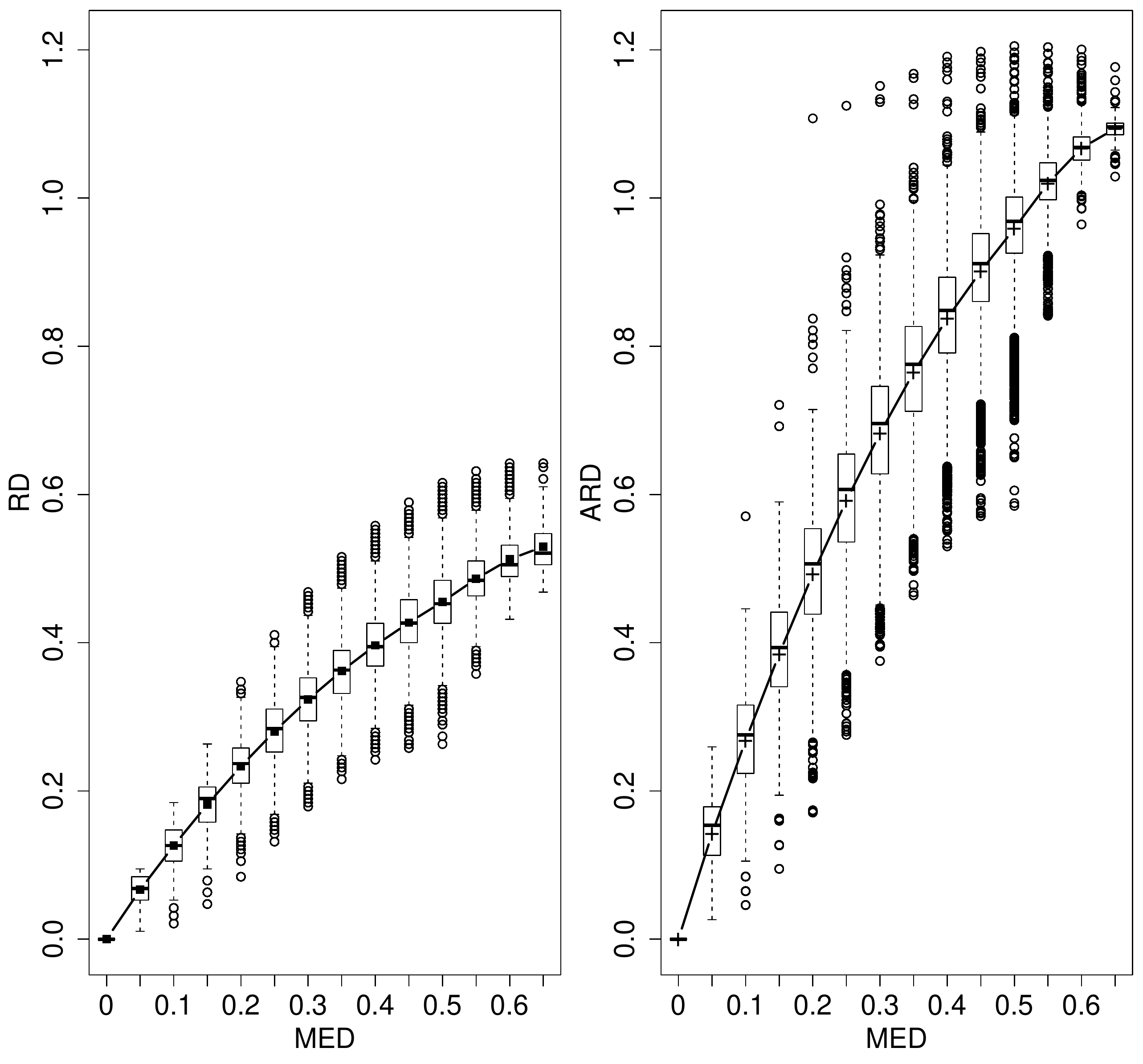}
\caption{RD (left) and ARD (right) versus MED for $r=s=3$ and $n=20$. The boxplots represent the conditional distributions for a given value of MED, and the curves depict the conditional means.}
\label{fig:aRDvsMED}
\end{figure}

For $r=s=3$ and $n=20$ the class of all possible confusion matrices is considerably larger, namely $|\mathcal N(3,3,20)|=2806281$, but still not prohibitive, so its exhaustive enumeration is yet feasible. Therefore, the ARD, the MED and the RD were obtained for each of these possible confusion matrix configurations, yielding a large amount of interesting information. Figure \ref{fig:aRDvsMED} shows boxplots for the conditional distributions of the RD (left) and the ARD (right), given the MED, along with the (mean) regression curve. These plots contain the same information as Figure \ref{fig:MED_RD}, but a first notable difference is that now the RD corresponding to a given MED is no longer a single value, as it happened for $r=s=2$; instead, for $r=s=3$ all the possible confusion matrices with the same MED result in a wide range of different RD values.


\begin{figure}[h!t]\centering
\includegraphics[width=.4\textwidth]{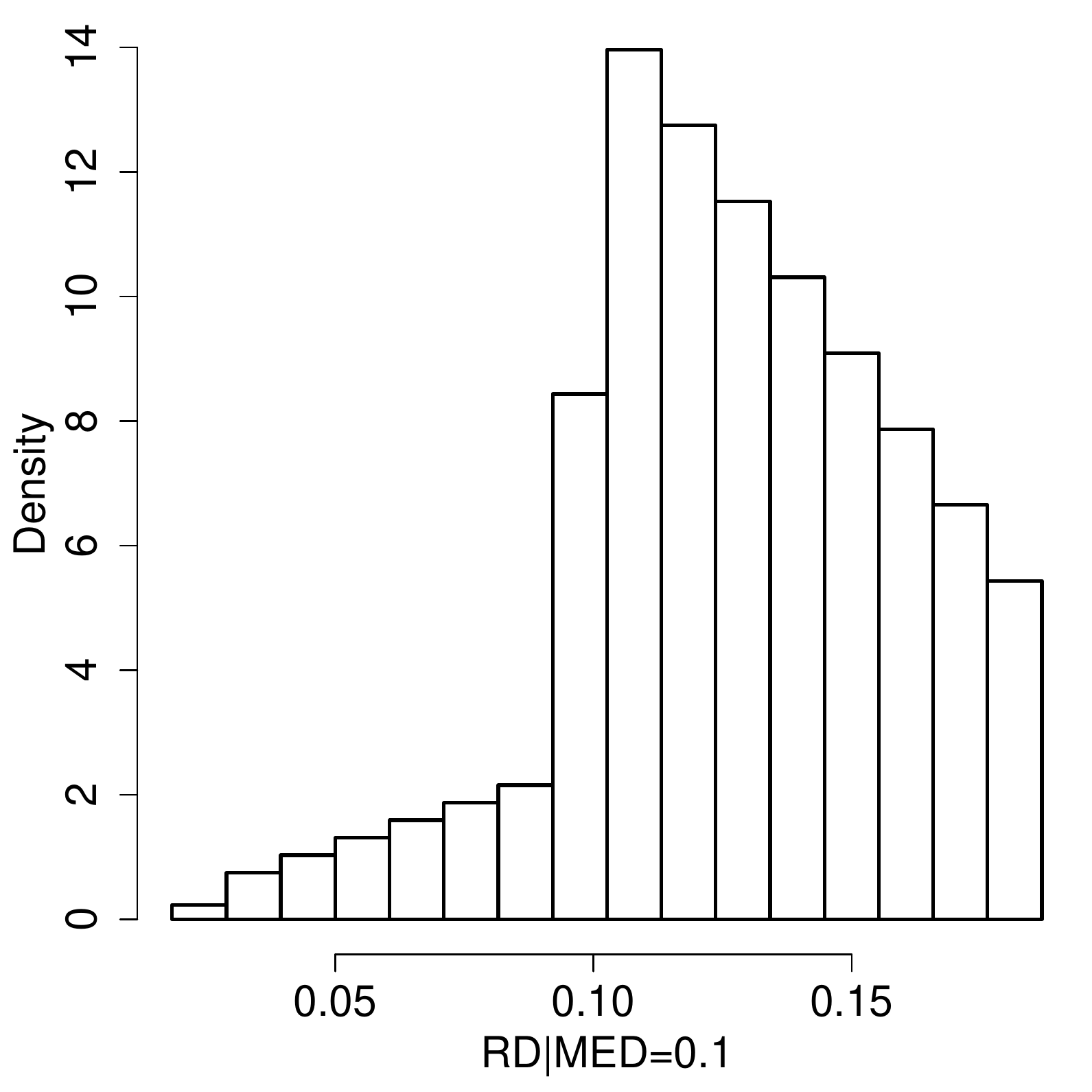}\hspace{0.1\textwidth}\includegraphics[width=.4\textwidth]{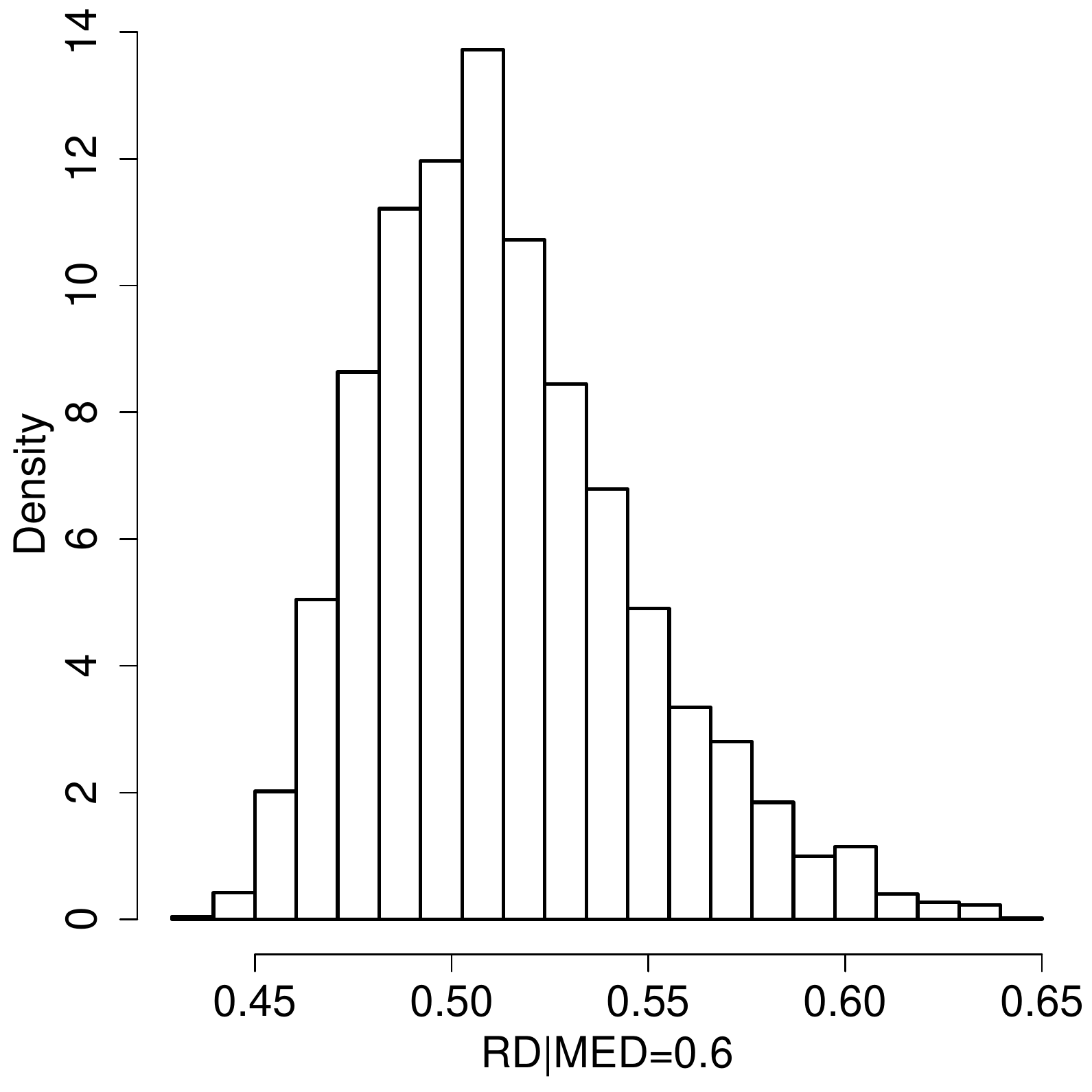}
\caption{Conditional distribution of the RD given ${\rm MED}=0.1$ (left) and ${\rm MED}=0.6$ (right) for $r=s=3$ and $n=20$.}
\label{fig:RDMED0106}
\end{figure}

Most of the conditional distributions given the MED are fairly symmetric, but it is worth remarking some interesting features that arise, especially, for very low or very high MED values. For instance, Figure \ref{fig:RDMED0106} focuses on the distribution of the RD given the particular values of ${\rm MED}=0.1$ (left) and ${\rm MED}=0.6$ (right), which clearly show a high degree of skewness. The conditional distribution of the ARD also shows some peculiarities: its maximum value (${\rm ARD}=1.205$) is attained for a confusion matrix with ${\rm MED}=0.5$, but its conditional mean attains its maximum at ${\rm MED}=0.65$ (the maximum possible MED value). The outlier in the conditional distribution of the ARD given ${\rm MED}=0.2$ is particularly striking, with a value of $1.108$, attained at the confusion matrix
$${\mathbf N}=\begin{pmatrix}16&1&1\\1&0&0\\1&0&0\end{pmatrix},$$
despite it is only at 4 data point transfers from the perfect agreement situation.

\begin{figure}\centering
\includegraphics[width=.4\textwidth]{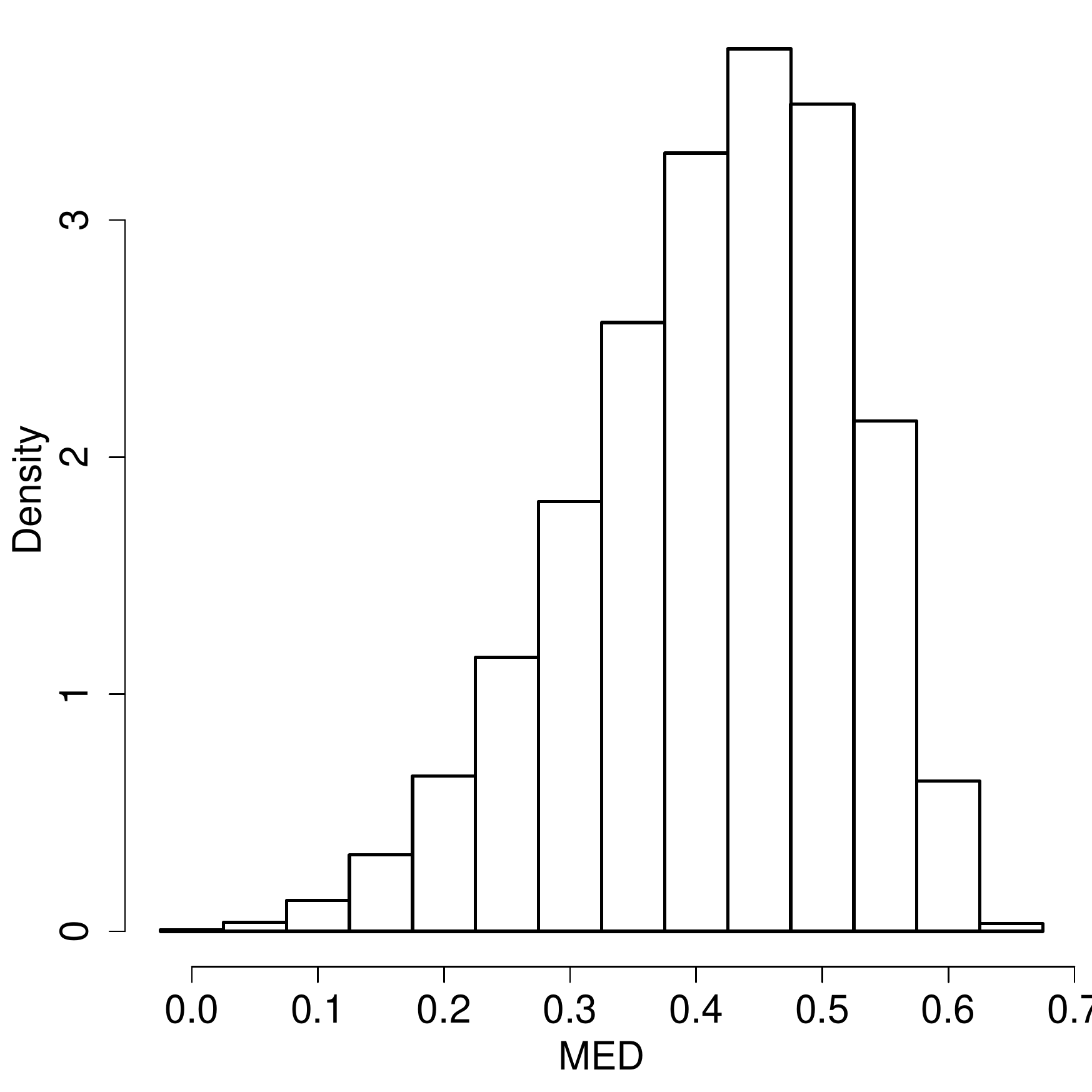}
\caption{Distribution of the possible MED values for $r=s=3$ and $n=20$.}
\label{fig:MEDdist}
\end{figure}

This extensive enumeration study is also useful to inspect the individual distributions of each criterion. For example, Figure \ref{fig:MEDdist} shows the distribution of all the MED values for $r=s=3$ and $n=20$, and reveals a very different scenario from the one that Figure 1 in \cite{St04} suggested. In Steinley's simulation study, the distribution of the MED appeared to be somewhat uniform, which strongly contrasts with the distribution shape shown in Figure \ref{fig:MEDdist} from exhaustive enumeration. The reason is that in Steinley's paper the distribution of the MED is investigated by aggregation of all the multiple simulation conditions. And, as noted before, these simulation conditions involved uniformly varying the DO from $0.05$ to $0.95$. It was already noted above that the DO is not exactly the same as the MED, but they are closely related, so forcing a fixed given number of simulations for every DO level naturally results in a (nearly) uniform distribution for the MED. However, the exhaustive inspection of all the possible confusion matrix configurations in Figure \ref{fig:MEDdist} shows that the MED distribution is quite different from the uniform one.

For higher values of  $r$, $s$ or $n$, it is not possible to enumerate all the confusion matrices in $\mathcal N(r,s,n)$ anymore, since its cardinality becomes exorbitant. An alternative way to approximate the criteria distributions, for these higher values of $r$, $s$ and $n$, would be to randomly sample a large number of matrices from $\mathcal N(r,s,n)$ (in an equiprobable way), and then compute their MEDs, RDs and ARDs in order to obtain an equivalent approximation of Figure \ref{fig:aRDvsMED}. This suggestion is not without problems, either, for two reasons: first, it is not straightforward to uniformly sample from $\mathcal N(r,s,n)$, see Appendix C for a valid procedure; and second, the fact that some of the possible MED values occur only for a low number of confusion matrices makes it difficult to procure an accurate approximation of the conditional distributions given such MED values. For example, from the exhaustive enumeration of $\mathcal N(3,3,20)$ it follows that the probability of obtaining a confusion matrix with ${\rm MED}=0.65$ by uniform sampling is approximately $1.6\times 10^{-3}$, so a large simulation size would be required in order to approximate the distribution of the other distances given ${\rm MED}=0.65$.

\begin{figure}\centering
\includegraphics[width=.4\textwidth]{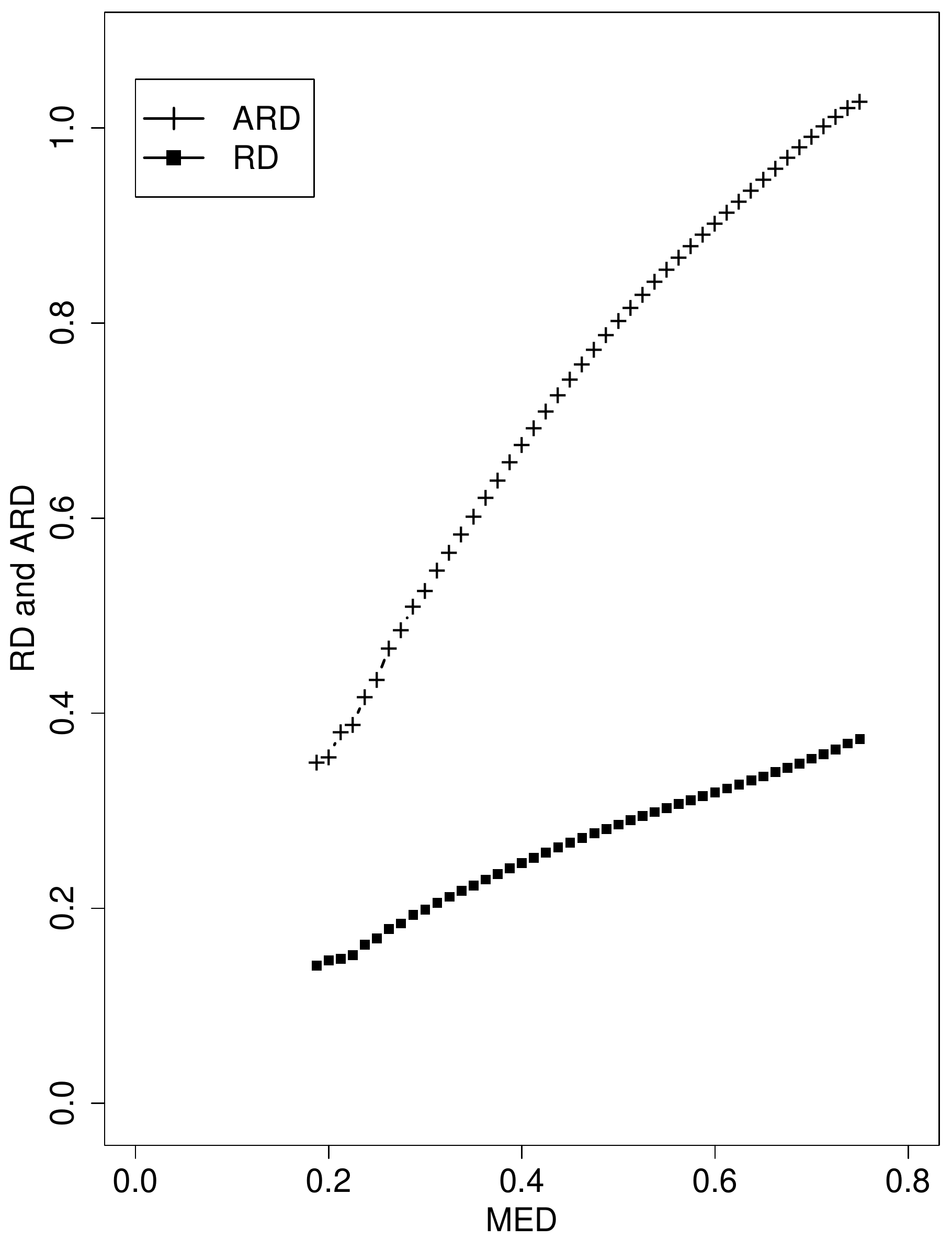}\hspace{.05\textwidth}\includegraphics[width=.4\textwidth]{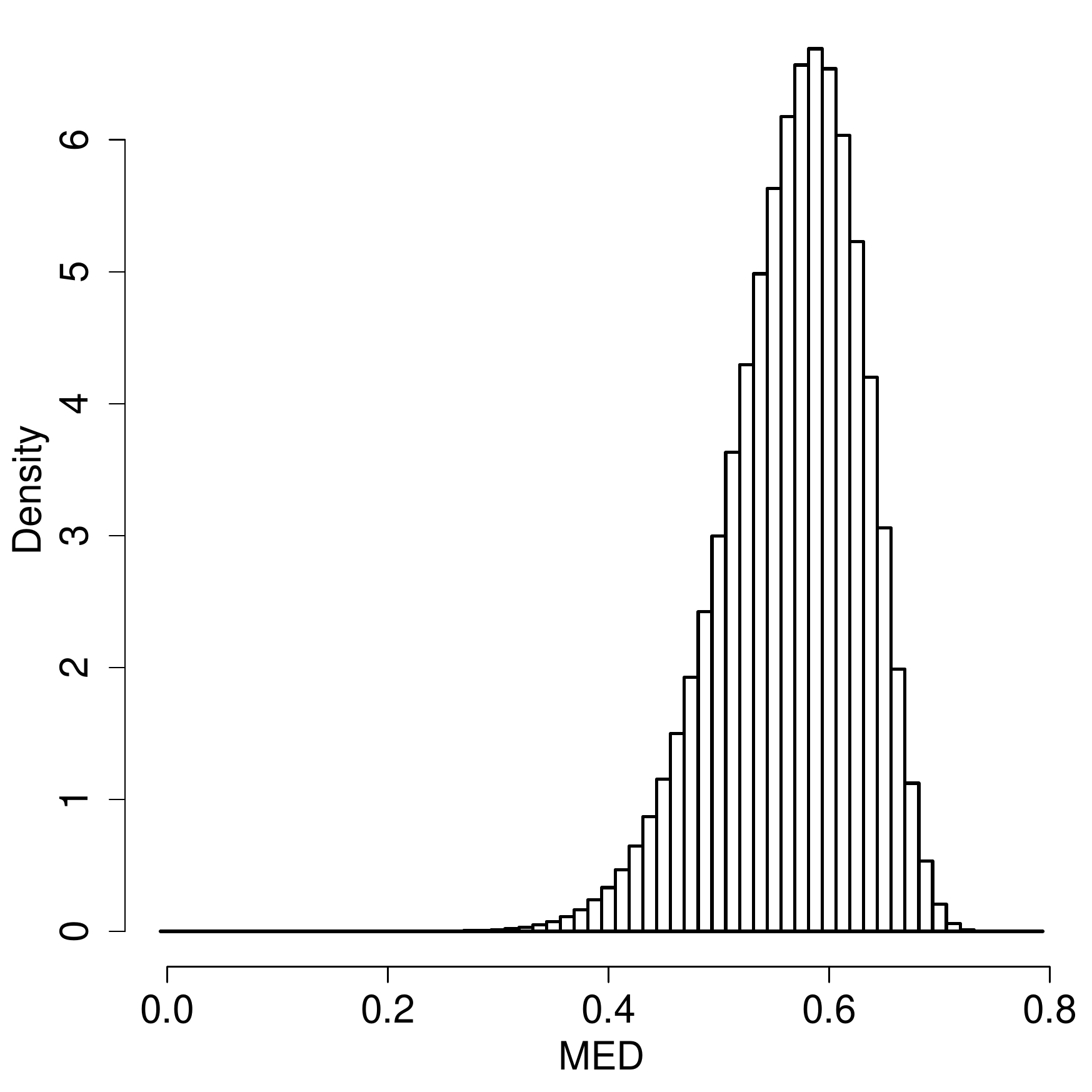}
\caption{Approximated conditional means of the RD and the ARD given the MED (left) and distribution of possible MED values (right), for $r=s=5$ and $n=80$.}
\label{fig:dist2}
\end{figure}

In any case, following the procedure suggested in Appendix C, a random sample of size $10^7$ was drawn from $\mathcal N(5,5,80)$, and the values of the MED, RD and ARD for these confusion matrices were recorded. It must be remarked that the cardinality of $\mathcal N(5,5,80)$ is approximately $2.309\times10^{23}$, which makes the exhaustive enumeration approach unfeasible. From that sample, it is possible to approximate the conditional means of the RD and ARD given the MED (Figure \ref{fig:dist2}, left) and to provide an approximate analogue of Figure \ref{fig:MEDdist} for $n=80$ and $r=s=5$ (Figure \ref{fig:dist2}, right). Notice also that, even if the possible MED values are $\{i/80\colon i=0,1,\dots,64\}$ (because $\max{\rm MED}=0.8$ for $r=s=5$ and $n=80$), the range of MED values for which 10 or more observations were obtained in this particular sample reduced to $\{i/80\colon i=15,16,\dots,60\}$ (i.e., the others had sample frequencies smaller than $10^{-6}$), and that is why Figure \ref{fig:dist2} presents some missing parts.

\section{Discussion}\label{sec:6}

Regarding the task of comparing two partitions of a finite data set, surely the confusion matrix is the object that yields the most complete information. However, when it has a considerable number of cells, it provides somehow too many details and it becomes necessary to resort to some summary statistic to extract useful information. The MED, the RD and the ARD are examples of such summary statistics, each of them offering a different synopsis.

The first two represent empirical versions of distances between whole-space clusterings and, intuitively, correspond to computing the proportion of ``differently placed'' individual data points (in case of the MED) or data pairs (for the RD) along the compared clusterings. Considering data pairs instead of individual data points appears somehow less intuitive. In fact, as pointed out in \citet[Section 4]{HA85}, one could equally consider using data triplets or, more generally, data $k$-tuples. Comparisons, however, become more and more intricate as $k$ increases. It is in this sense that the choice of $k=1$ (i.e., the MED) represents the simplest option.

But it is not just a matter of simplicity. As long as $\min\{r,s\}>2$, the RD also shows a more serious and undesirable drawback: the case of completely unrelated clusterings does not correspond to the most dissimilar clustering pair, according to the RD, and this phenomenon becomes more and more severe as the clustering sizes increase (as shown in Figure \ref{fig:NRDi}). 
This unfortunate feature is not shared by the MED, which does point out unrelated clusterings as an instance of extreme dissimilarity, and furthermore shows a maximum value that quickly approaches 1 as $\max\{r,s\}$ increases.

A possibility to correct the aforementioned flaw is to consider the relative size of the RD with respect to the average RD value when the two clusterings are generated at random, this is what the ARD provides. 
This exhibits the natural advantage of creating a criterion that is always centred at 1 for the null case, but on the other hand introduces a distorting element that further complicates the interpretation: now the ARD represents the relative size of the proportion of differently treated data pairs in the compared clusterings with respect to a baseline, taken as the average value of that proportion when the cluster labels are assigned at random while maintaining the number of clusters and cluster sizes fixed. This also implies that, if the baseline changes (as usually happens when inspecting two different confusion matrices), then the relative comparison of the two scenarios by means of ARD scores becomes unclear.

An alternative remedy, also aimed to examine the relative size of a criterion, but this time against the worst possible case, is to normalize such a criterion with respect to its maximum value. This is a different kind of adjustment, which does not produce a criterion that is centred at 1 for a null model (in fact, it does not rely on a specific null model), however it ensures that all the resulting values lie on $[0,1]$ instead. When applied to the MED and the RD it results in the new NMED and NRD criteria, which are not advised to be used alone, but jointly with their unnormalized counterparts, since the latter retain the most straightforward interpretation. 
In addition, not achieving its maximum for unrelated clusterings also hinders this approach for the RD, as shown in Figures \ref{fig:null100} and \ref{fig:null400}, since it entails that the distribution of the NRD can be far from 1 under the null model. In contrast, the NMED distribution is indeed close to its upper bound of 1 in the null case, more so for higher sample sizes, although it must be pointed out that it seems more and more unlikely to reach this upper bound as the number of clusters increase.

In any case, it seems clear that the study of the distributions of all these criteria (the MED, the RD and the ARD) deserves further consideration, since its investigation through exhaustive enumeration or uniform sampling from the set of all possible confusion matrices has revealed some previous misconceptions and unexpected features.

\bigskip

\noindent{\bf Acknowledgments.}  The author acknowledges the support of the Spanish Ministerio de Econom\'\i a y Competitividad grant MTM2016-78751-P and the Junta de Extremadura grant GR18016.

\appendix

\section*{Appendix A: R function for misclassification error distance computation}

Recall from Section \ref{sec:comp} that, given a confusion matrix ${\mathbf N}=(n_{ij})\in\mathcal N(r,s,n)$ with $r\leq s$, the main computational problem is to find
\begin{equation}\label{eq:lsap}
\min_{\sigma\in\mathcal P_s}\sum_{i=1}^sm_{i,\sigma(i)},
\end{equation}
where $\mathcal P_s$ denotes the set of all possible permutations of $s$ elements. Here, $n_{i+}=n_{ij}=0$ for all $i=r+1,\dots,s$ (if any) and $m_{ij}=n_{i+}+n_{+j}-2n_{ij}$ for $i,j=1,\dots,s$. Fortunately, (\ref{eq:lsap}) is a linear sum assignment problem, whose solution can be efficiently found through the function {\tt solve\_LSAP} included in the R library {\tt clue} \citep{Ho05,Ho18}. So once that library is loaded, with the command {\tt library(clue)}, a function to compute the MED from two equal-size vectors containing the cluster labels with respect to each clustering can be obtained through the following simple code:

\begin{verbatim}
med <- function(labels1, labels2){
  n <- length(labels1)
  N <- table(labels1, labels2)
  r <- nrow(N)
  s <- ncol(N)
  if (r>s) N <- t(N); r <- nrow(N); s <- ncol(N)
  if (r<s) N <- rbind(N, matrix(0, nrow = s-r, ncol = s))
  M <- matrix(rowSums(N), nrow = s, ncol = s) +
    matrix(colSums(N), nrow = s, ncol = s, byrow = TRUE) - 2 * N
  optimal.permutation <- solve_LSAP(M)
  result <- sum(M[cbind(seq_along(optimal.permutation),
    optimal.permutation)]) / (2 * n)
  return(result)
}
\end{verbatim}

\section*{Appendix B: The maximum Rand distance}

Assuming as true the conjecture that there is always a maximizer of the RD of the form (\ref{eq:argmaxRD}), here it is shown that the maximum value of the RD satisfies (\ref{eq:maxRD}). First notice that, for a confusion matrix of the form (\ref{eq:argmaxRD}), the function $M(q_1,\dots,q_s)=n(n-1){\rm RD}=n^2\widehat d_{\rm H}$ can be explicitly written as
\begin{equation}\label{eq:M}
M(q_1,\dots,q_s)=2(r-1)q_1-\sum_{j=1}^sq_j^2+c,
\end{equation}
where $c=(n-r+1)^2+(r-1)(r-2)$. Then, the goal is to maximize $M(q_1,\dots,q_s)$ with the constraint that the total number of data points is $n$, which for the matrix (\ref{eq:argmaxRD}) yields $\sum_{j=1}^sq_j+r-1=n$. The method of Lagrange multipliers yields the maximizer over real-valued choices of $q_1,\dots,q_s$ as $q_1^*=\{n-2(r-1)\}/s+r-1$, $q_2^*=\cdots=q_s^*=\{n-2(r-1)\}/s$, but recall that the goal is to find the maximizer for nonnegative integer values of $q_1,\dots,q_s$.

As in Section \ref{sec:arbitrary}, write $n-2(r-1)=ks+\ell$, with $k\in\mathbb N$ and $\ell\in\{0,1,\dots,s-1\}$. If $\{n-2(r-1)\}/s\in\mathbb N$ (corresponding to the case $\ell=0$), then the real-valued maximizer is also integer-valued and leads to the maximizer and maximum value announced in (\ref{eq:argmaxRD}) and (\ref{eq:maxRD}) for $\ell=0$.

If $\ell>0$ then the real-valued maximizer is rational, with all the coordinates $q_1^*,\dots,q_s^*$ having the same fractional part $\ell/s$. Due to the total sum constraint, to find the integer-valued maximizer of (\ref{eq:M}) these fractional reminders need to be re-distributed into the coordinates $q_1,\dots,q_s$, to make them integer, while at the same time trying to decrease the value of $M(q_1,\dots,q_s)$ as less as possible with respect to the real-valued maximizer. To achieve this, first note that (\ref{eq:M}) is a concave function with the same curvature along every direction, so the least decrease with integer coordinates with respect to the real-valued maximum corresponds to rounding up to the least greater integer as few coordinates of $q_1^*,\dots,q_s^*$ as possible. Having $s$ fractional reminders of size $\ell/s$, that entails that the integer-valued maximizer is found by rounding up exactly $\ell$ coordinates to the least greater integer (and rounding down the remaining $s-\ell$ coordinates). Finally, since $q_2,\dots,q_s$ play a symmetric role in (\ref{eq:M}), the only two cases to study comprise, either to round up $q_1^*$ and $\ell-1$ of the remaining coordinates, or to round up $\ell$ coordinates among $q_2^*,\dots,q_s^*$ (say, the first $\ell$ of them). The first of these cases yields $q_1=k+r$, $q_2=\cdots=q_\ell=k+1$, $q_{\ell+1}=\dots=q_s=k$, while the second one entails $q_1=k+r-1$, $q_2=\cdots=q_{\ell+1}=k+1$, $q_{\ell+2}=\dots=q_s=k$. It is easy to check that these two choices achieve the same value for $M(q_1,\dots,q_s)$, and the second one agrees with the form posited in (\ref{eq:argmaxRD}) and (\ref{eq:maxRD}), which is, thus, valid for arbitrary $\ell$.

\section*{Appendix C: Uniform sampling from $\mathcal N(r,s,n)$}

A composition of a positive integer $n$ into $k$ parts is a representation $n=\sum_{i=1}^kp_i$ in which $p_1,\dots,p_k$ are non-negative integers and the order of the summands matters. There are $J(n,k)={n+k-1\choose n}$ possible compositions of $n$ into $k$ parts, and it is easy to draw a composition at random (uniformly) without necessarily generating the set of all of them \citep[see][Chapters 5 and 6]{NW78}.

The entries of any confusion matrix ${\mathbf N}\in\mathcal N(r,s,n)$ constitute a composition of $n$ into $r\cdot s$ parts. The additional conditions $n_{i+}>0$ and $n_{+j}>0$ for all $i,j$, imposed in the definition of $\mathcal N(r,s,n)$ to ensure that the sizes of the associated compared clusterings match $r$ and $s$, respectively, can be checked after arranging each drawn composition of $n$ into $r\cdot s$ parts by columns (say) into an $r\times s$ matrix, so uniform sampling from $\mathcal N(r,s,n)$ is guaranteed by rejection sampling.

For the simulations in Section \ref{sec:diverging} a sample of size $10^7$ was drawn from $N(5,5,80)$ using the previous approach. The recorded rejection rate during the process was very low, approximately $0.47\%$, so the sampling algorithm is very efficient. Moreover, that rejection rate can also be interpreted as an estimate of the proportion of compositions of $n=80$ into $r\cdot s=25$ parts that cannot be converted (by columns) into a confusion matrix of $\mathcal N(5,5,80)$ and, since $J(80,25)\simeq 2.319\times 10^{23}$, that yields an estimate of $2.309\times 10^{23}$ for the cardinality of $\mathcal N(5,5,80)$.

\bibliographystyle{apalike}

\begin{thebibliography}{}

\bibitem[{Aghaeepour {\it et al.}}, 2013]{Aal13} Aghaeepour, N., Finak,	G., The FlowCAP Consortium,	The DREAM Consortium, Hoos,	H., Mosmann, T.R., Brinkman, R., Gottardo, R. and Scheuermann, R.H. (2013). Critical assessment of automated flow cytometry analysis techniques. {\it Nature Methods}, {\bf 10}, 228--238.


\bibitem[Anderson, 1935]{A35} Anderson, E. (1935). The irises of the Gaspe Peninsula. {\it Bulletin of the American Iris Society}, {\bf 59}, 2--5.

%

\bibitem[{Azizyan {\it et al.}}, 2015]{A15} Azizyan, M., Chen, Y.-C., Singh, A. and Wasserman, L. (2015). Risk bounds for mode clustering. \emph{arXiv:1505.00482}.

%
%
%

\bibitem[{Ben-David, von Luxburg and P\'al}, 2006]{Bal06} Ben-David, S., von Luxburg, U. and P\'al,
    D. (2006). A sober look at clustering stability. In G. Lugosi and H.-U. Simon, editors, {\it
    Proceedings of the 19th Annual Conference on Learning Theory (COLT)}, pp. 5--19. Springer, Berlin.


\bibitem[{Burkard, Dell'Amico and Martello}, 2009]{BDM09} Burkard, R., Dell'Amico, M. and Martello, S. (2009) {\it Assignment Problems}. SIAM, Philadelphia.

%
%
%
%
%
%

\bibitem[Chac\'on, 2015]{Ch15} Chac\'on, J.E. (2015). A population background for
nonparametric density-based clustering. {\it Statistical Science}, {\bf 30}, 518--532.

\bibitem[Chac\'on, 2019]{Ch18} Chac\'on, J.E. (2019). Mixture model modal clustering. {\it Advances in Data Analysis and Classification}, {\bf 13}, 379--404.

%

\bibitem[{Charon {\it et al.}}, 2006]{CDGH06} Charon, I., Den{\oe}ud, L., Gu\'enoche, A. and Hudry, O. (2006). Maximum transfer distance between partitions. {\it Journal of Classification}, {\bf 23}, 103--121.

\bibitem[{Charon, Den{\oe}ud and Hudry}, 2007]{CDH07} Charon I., Den{\oe}ud L., Hudry O. (2007). Maximum de la distance de transfert \`{a} une partition donn\'{e}e. {\it Math\'{e}matiques et Sciences Humaines}, {\bf 179}, 45--83.

%
%

\bibitem[Day, 1981]{D81} Day, W.H.E. (1981). The complexity of computing metric distances between
    partitions. {\it Mathematical Social Sciences}, {\bf 1}, 269--287.


\bibitem[Den{\oe}ud, 2008]{D08} Den{\oe}ud, L. (2008) Transfer distance between partitions. {\it Advances in Data Analysis and Classification}, {\bf 2}, 279--294.

\bibitem[Den{\oe}ud and Gu\'enoche, 2006]{DG06} Den{\oe}ud, L. and Gu\'enoche, A. (2006). Comparison of distance indices between
partitions. In V. Batagelj, H.-H. Bock, A. Ferligoj and A. \v{Z}iberna (Eds.), {\it Data Science and Classification}, 21--28. Springer-Verlag, Berlin.

%
%
%
%
%

\bibitem[Filkov and Skiena, 2004]{FS03} Filkov, V. and Skiena, S. (2004). Integrating microarray data by consensus clustering. {\it International Journal on Artificial Intelligence Tools}, {\bf 13}, 863--880.

%

\bibitem[Fowlkes and Mallows, 1983]{FM83} Fowlkes, E. B. and Mallows, C. L. (1983). A method for comparing two hierarchical clusterings. {\it Journal of the American Statistical Association}, {\bf 78}, 553--569.

\bibitem[Fraley and Raftery, 2002]{FR02} Fraley, C. and Raftery, A.E. (2002). Model-based clustering, discriminant analysis, and density estimation. {\it Journal of the American Statistical Association}, {\bf 97}, 611--631.

%
%

\bibitem[Gart, 1970]{G70} Gart, J. J. (1970). A locally most powerful test for the symmetric folded binomial distribution. {\it Biometrics}, {\bf 26}, 129--138.

\bibitem[Gates and Ahn, 2017]{GA17} Gates, A. J. and Ahn, Y.-Y. (2017). The impact of random models on clustering similarity. {\it Journal of Machine Learning Research}, {\bf 18}, 3049--3076.

\bibitem[Goodman and Kruskal, 1979]{GK79} Goodman, L. A. and Kruskal, W. H. (1979). {\it Measures of Association
for Cross Classifications}. Springer-Verlag, New York.

\bibitem[Gusfield, 2002]{G02} Gusfield, D. (2002). Partition-distance: A problem and class of perfect graphs arising in clustering. {\it Information Processing Letters}, {\bf 82}, 159--164.

\bibitem[{Gy\"{o}rfi {\it et al.}}, 2002]{Gal02} Gy\"{o}rfi, L., Kohler, M., Krzy\.{z}ak, A. and Walk, H. (2002). {\it A Distribution-Free Theory of Nonparametric Regression}. Springer-Verlag, New York.

%
%

\bibitem[Hennig, 2019]{H19} Hennig, C. (2019). Cluster validation by measurement of clustering characteristics relevant to the user. In C. H. Skiadas and J. R. Bozeman (Eds.), {\it Data Analysis and Applications 1: Clustering and Regression, Modeling-estimating, Forecasting and Data Mining}, 1--24. ISTE Ltd, London.

\bibitem[Hornik, 2005]{Ho05} Hornik, K. (2005). A CLUE for CLUster Ensembles. {\it Journal of Statistical Software}, {\bf 14}(12).

\bibitem[Hornik, 2018]{Ho18} Hornik, K. (2018). {\it clue: Cluster ensembles}. R package version 0.3-55.

\bibitem[Hubert and Arabie, 1985]{HA85} Hubert, L. and Arabie, P. (1985). Comparing partitions. {\it Journal of Classification}, {\bf 2}, 193--218.

\bibitem[Jain and Dubes, 1988]{JD88} Jain, A. K. and Dubes, R. C. (1988). {\it Algorithms for Clustering Data}. Prentice-Hall, Inc., Englewood Cliffs.

\bibitem[Klemel\"{a}, 2009]{K09} Klemel\"{a}, J. (2009). {\it Smoothing of Multivariate Data: Density Estimation and Visualization}. John Wiley \& Sons, Inc., Hoboken.

\bibitem[Lee, 1990]{L90} Lee, A. J. (1990). {\it $U$-Statistics: Theory and Practice}. Marcel Dekker, Inc., New York.

%
%
%
%
%
%

\bibitem[Meil\u{a}, 2005]{M05} Meil\u{a}, M. (2005) Comparing clusterings---an axiomatic view. In S. Wrobel and L. De Raedt, editors, {\it Proceedings of the International Machine Learning Conference (ICML)}. ACM Press, New York.

\bibitem[Meil\u{a}, 2007]{M07} Meil\u{a}, M. (2007) Comparing clusterings---an information based distance. {\it Journal of Multivariate Analysis}, {\bf 98}, 873 -- 895.

\bibitem[Meil\u{a}, 2012]{M12} Meil\u{a}, M. (2012) Local equivalences of distances between
clusterings---a geometric perspective. {\it Machine Learning}, {\bf 86},  369--389.

\bibitem[Meil\u{a}, 2016]{M16} Meil\u{a}, M. (2016). Criteria for comparing clusterings. In C. Hennig, M. Meil\u{a}, F. Murtagh and R. Rocci (Eds.), {\it Handbook of Cluster Analysis}, 619--635. CRC Press, Boca Raton.

\bibitem[Milligan, 1996]{M96} Milligan, G. W. (1996). Clustering validation: results and implications for applied analyses. In P. Arabie, L. J. Hubert, and G. De Soete (Eds.), {\it Clustering and Classification}, 341--375. World Scientific Publishing, River Edge.

\bibitem[Mirkin, 1996]{Mir96} Mirkin, B. G. (1996). {\it Mathematical Classification and Clustering}. Kluwer Academic Press, Dordrecht.

\bibitem[Mirkin and Chernyi, 1970]{MC70} Mirkin, B. G. and Cherny, L. B. (1970). On a distance measure between partitions
of a finite set. {\it Automation and Remote Control}, {\bf 31}, 786--792.

\bibitem[Milligan and Cooper, 1986]{MC86} Milligan, G. W. and Cooper, M. C. (1986). A study of the comparability of external criteria for
hierarchical cluster analysis. {\it Multivariate Behavioral Research}, {\bf 21}, 441--458.

\bibitem[Nijenhuis and Wilf, 1978]{NW78} Nijenhuis, A. and Wilf, H. S. (1978). {\it Combinatorial Algorithms. For Computers and Calculators} (2nd ed.). Academic Press, New York.

\bibitem[Papadimitriou and Steiglitz, 1982]{PS82} Papadimitriou, C. and Steiglitz, K. (1982). {\it Combinatorial Optimization: Algorithms and Complexity}. Prentice Hall, Englewood Cliffs.

%

\bibitem[R Core Team, 2019]{R18} R Core Team (2019). {\it R: A Language and Environment for Statistical Computing}. R Foundation for Statistical Computing, Vienna, Austria.

\bibitem[Rand, 1971]{R71} Rand, W. M. (1971). Objective criteria for the evaluation of clustering methods. {\it Journal of the American
Statistical Association}, {\bf 66}, 846--850.

%
%

\bibitem[R\'egnier, 1965]{R65} R\'egnier, S. (1965) Quelques aspects math\'ematiques des
    probl\`emes de classification automatique. {\it I.C.C. Bulletin}, {\bf 4}, 175--191. Reprint (1983) in {\it Math\'ematiques et Sciences
    Humaines}, {\bf 82}, 31--44.

%

\bibitem[Rossi, 2015]{R15} Rossi, G. (2015). Hamming distance between partitions,
clustering comparison and information. {\it Proceedings of the International Conference on Pure Mathematics, Applied Mathematics and Computational Methods (PMAMCM2015)}, 101--107.

%

\bibitem[Steinley, 2003]{St03} Steinley, D. (2003). Local optima in $K$-means clustering: What you don't know
may hurt you. {\it Psychological Methods}, {\bf 8}, 294--304.

\bibitem[Steinley, 2004]{St04} Steinley, D. (2004). Properties of the Hubert--Arabie adjusted Rand index. {\it Psychological Methods}, {\bf 9}, 386--396.

\bibitem[Steinley and Brusco, 2018]{SB18} Steinley, D. and Brusco, M. J. (2018). A note on the expected value of the Rand index. {\it British Journal of Mathematical and Statistical Psychology}, {\bf 71}, 287--299.

\bibitem[{Steinley, Brusco and Hubert}, 2016]{SBH16} Steinley, D., Brusco, M. J. and Hubert, L. (2016). The variance of the adjusted Rand index. {\it Psychological Methods}, {\bf 21}, 261--272.


\bibitem[{Van Mechelen {\it et al.}}, 2018]{VMal18} Van Mechelen, I., Boulesteix, A.-L., Dangl, R., Dean, N., Guyon, I., Hennig, C., Leisch, F. and Steinley, D. (2018). Benchmarking in cluster analysis: A white paper. {\tt arXiv:1809.10496v2}.

\bibitem[von Luxburg and Ben-David, 2005]{vLBD05} von Luxburg, U. and Ben-David, S. (2005). Towards a statistical theory for clustering. In {\it PASCAL workshop on Statistics and Optimization of Clustering}.

\bibitem[von Luxburg, 2010]{vL10} von Luxburg, U. (2010). Clustering stability: an overview. {\it Foundations and Trends in Machine Learning}, {\bf 2}, 235--274.

\bibitem[Wallace, 1983]{W83} Wallace, D. L. (1983). A method for comparing two hierarchical clusterings: comment. {\it Journal of the American Statistical Association}, {\bf 78}, 569--576.

%

\bibitem[Warrens, 2008]{W08} Warrens, M. J. (2008). On the equivalence of Cohen's kappa and the Hubert-Arabie adjusted Rand index. {\it Journal of Classification}, {\bf 25}, 177--183.

\bibitem[Wasserman, 2018]{W18} Wasserman, L. (2018). Topological data analysis. {\it Annual Review of Statistics and Its Application}, {\bf 5},  501-532.

\end{thebibliography}

\end{document}